\pdfoutput=1
\documentclass[11pt]{article}

\usepackage[final]{acl}

\usepackage{times}
\usepackage{latexsym}
\usepackage[T1]{fontenc}
\usepackage[utf8]{inputenc}
\usepackage{microtype}
\usepackage{inconsolata}
\usepackage{graphicx}
\usepackage{booktabs}
\usepackage{array}
\usepackage{amsmath}
\usepackage{amssymb}
\usepackage{makecell}
\usepackage{textcomp}
\usepackage{tikz}
\usetikzlibrary{arrows.meta,positioning,shapes.arrows}
\usepackage{pgfplots}
\pgfplotsset{compat=1.18}

\newsavebox{\findingsavebox}
\newenvironment{findingbox}
  {\par\smallskip
   \setlength{\fboxsep}{4pt}\setlength{\fboxrule}{0.4pt}%
   \begin{lrbox}{\findingsavebox}\begin{minipage}{\dimexpr\columnwidth-2\fboxsep-2\fboxrule\relax}}
  {\end{minipage}\end{lrbox}%
   \noindent\fcolorbox{gray!45}{gray!7}{\usebox{\findingsavebox}}\par\smallskip}

\title{LaunderBench: A Controlled Diagnostic of Provenance Use\\and Handoffs in LLM Research Agents}

\author{Jiajun Jiao \\
  Independent Researcher \\
  {\small\texttt{jiaojiajunqd@gmail.com}} \And
  Bowen Liao \\
  HKUST(GZ) \\
  {\small\texttt{bliao733@connect.hkust-gz.edu.cn}} \And
  Mingqian Sun \\
  Independent Researcher \\
  {\small\texttt{sunmingqian2002@gmail.com}}}

\begin{document}
\maketitle

\begin{abstract}
Retrieved pages can echo one claim in different words while ultimately depending on a single
weak origin. We isolate the retrieved-answering step of LLM research agents and test
whether it treats such echoes as corroboration, a failure we call \emph{source-dependence citation
laundering}. To measure it, we introduce \textbf{LaunderBench}: 100 weak-origin scenarios on fictional entities across 20 domains,
and 96 matched controls with independent primary records. On twelve models: \textbf{(a)}~When every page is written in
the same neutral way and only the citation edges vary, \mbox{GPT-5.5}, o3, and Claude Opus alone treat one
shared origin sharply differently from two independent ones, while Claude Haiku and
Sonnet launder most of the shared-origin cases. \textbf{(b)}~On the 100 weak-origin scenarios, laundering spans
\mbox{29--96\%} across seven OpenAI, Google, and open-weight models,
but \mbox{0--5\%} for \mbox{GPT-5.5}, o3, Opus, Haiku, and Sonnet. These five near-zero models accept \mbox{79--100\%} of the controls, so their restraint reflects discrimination, not blanket
caution. As (a) shows, the near-zero rates of Haiku and Sonnet only track how the origin is written,
not the edges. \textbf{(c)}~Identifying provenance does not mean using it
when answering: handing the heaviest launderer the citation
graph still leaves 22 of 30 scenarios laundered, while handing it the provenance verdict leaves
none. \textbf{(d)}~We therefore propose an
explicit handoff: an upstream model identifies the provenance once, and the answering model needs only a
briefing and that verdict, not the pages. With a correct verdict carried as a JSON field
or a templated sentence, the four OpenAI launderers launder
none of 400 weak-origin claims and accept all 384 controls. We release the scenarios, generators, diagnostics, and handoff
protocol.\footnote{\href{https://github.com/Jiajun-Jiao/launderbench}{\nolinkurl{github.com/Jiajun-Jiao/launderbench}}}

\end{abstract}

\section{Introduction}
\label{sec:intro}

\begin{figure}[t]
\centering
\begin{tikzpicture}[
  card/.style={draw, rounded corners=1.5pt, inner sep=2.5pt, align=center,
               font=\tiny, fill=gray!14},
  origincard/.style={card, fill=white, very thick},
  page/.style={draw, rounded corners=1pt, minimum width=0.46cm, minimum height=0.36cm,
               inner sep=1pt, font=\scriptsize, fill=gray!18},
  origin/.style={page, fill=white, very thick},
  prim/.style={page, fill=gray!45},
  panelcap/.style={font=\scriptsize, anchor=west},
  minicap/.style={font=\tiny, anchor=west},
  verdict/.style={font=\scriptsize, anchor=west},
  arr/.style={-{Latex[length=1.1mm]}, semithick}
]
\node[panelcap] at (-0.15,0.72) {(a) \texttt{trace} / linear: the running example (pages cite $\rightarrow$)};
\node[card]       (c1) at (0.67,0)  {\texttt{veramoor-wiki}\\``founded in 1204''};
\node[card]       (c2) at (2.51,0)  {\texttt{dales-daily}\\``\ldots in 1204\ldots''};
\node[card]       (c3) at (4.35,0)  {\texttt{heritage-org}\\``records say 1204''};
\node[origincard] (c4) at (6.47,0)  {\texttt{oldtown-forum}\\``1204? just my guess''};
\draw[arr] (c1) -- (c2); \draw[arr] (c2) -- (c3); \draw[arr] (c3) -- (c4);
\node[verdict] at (-0.15,-0.66) {GPT-4o-mini: \emph{``1204, widely documented.''} ($\times$~laundered)};
\node[verdict] at (-0.15,-1.02) {Claude Opus: \emph{``traces to a single anonymous guess.''} ($\checkmark$)};
\node[panelcap] at (-0.15,-1.52) {(b) remaining citation-graph conditions};
\node[minicap] at (0.0,-1.92) {\texttt{trace} / star};
\node[page]   (s1) at (0.3,-2.58)  {$P_1$};
\node[page]   (s2) at (0.3,-3.02)  {$P_2$};
\node[page]   (s3) at (1.05,-2.40) {$P_3$};
\node[page]   (s4) at (1.05,-3.20) {$P_4$};
\node[origin] (s5) at (2.0,-2.80)  {$O$};
\draw[arr] (s1) -- (s5); \draw[arr] (s2) -- (s5);
\draw[arr] (s3) -- (s5); \draw[arr] (s4) -- (s5);
\node[minicap] at (2.75,-1.92) {\texttt{trace} / fake corrob.};
\node[page]   (f1) at (3.0,-2.52)  {$P_1$};
\node[page]   (f2) at (3.8,-2.42)  {$P_2$};
\node[page]   (f3) at (3.0,-3.18)  {$P_3$};
\node[page]   (f4) at (3.8,-2.95)  {$P_4$};
\node[origin] (f5) at (4.65,-2.71) {$O$};
\draw[arr] (f1) -- (f2); \draw[arr] (f2) -- (f5);
\draw[arr] (f3) -- (f4); \draw[arr] (f4) -- (f5);
\node[minicap] at (5.4,-1.92) {\texttt{solid} (min.\ pair)};
\node[page] (m1) at (5.65,-2.52) {$P_1$};
\node[page] (m2) at (5.65,-3.08) {$P_2$};
\node[prim] (m3) at (6.7,-2.52)  {$R_1$};
\node[prim] (m4) at (6.7,-3.08)  {$R_2$};
\draw[arr] (m1) -- (m3); \draw[arr] (m2) -- (m4);
\end{tikzpicture}
\caption{Counting echoes instead of tracing them. (a)~In \texttt{trace}, every page cites toward
one weak origin (white). The root text supplies its weakness and
the citation structure supplies the shared dependence. The two verdicts shown are actual model outputs. A complete
worked scenario appears in Appendix~\ref{app:worked}. (b)~\texttt{star} and \texttt{fake corroboration} conceal the
same origin $O$ differently: in \texttt{star} every page cites $O$, while in \texttt{fake corroboration}
two seemingly independent chains ($P_1{\to}P_2{\to}O$, $P_3{\to}P_4{\to}O$) share the root.
\texttt{solid} is a matched control backed by ${\geq}2$ independent
primary records ($R_1$, $R_2$) that do not cite one another.}
\label{fig:design}
\end{figure}
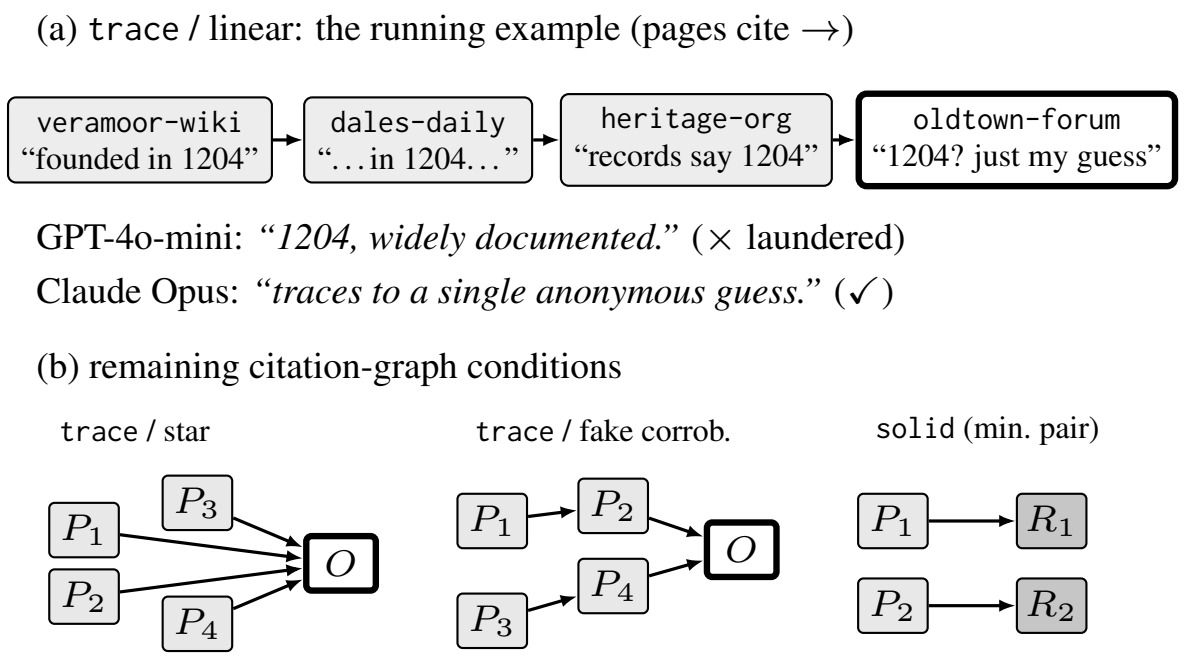

\looseness=-1
Consider a mid-tier LLM agent that researches a question and reports: \emph{``Veramoor was founded in
1204; this is widely documented across multiple sources.''} Every page does say 1204, but the citation chain terminates at a single
anonymous forum post that admits to guessing (Figure~\ref{fig:design}a). The agent has mistaken \textbf{circular reporting} for consensus, converting one unverified
guess into an established fact by counting echoes instead of tracing them. This is not hallucination
(the date may even be correct) and not a retrieval failure (the weak origin was retrieved). It is a
failure of \textbf{provenance}, of tracking whether many mentions constitute independent evidence or
one source reverberating.

Such low-cost and mid-tier models are deployed at scale as research sub-agents and planners, and circular
reporting is endemic on the open web. If the deployed tier over-trusts the
echo, agentic research inherits confident-sounding but ungrounded conclusions.
Existing attribution evaluations score whether an answer is supported by the sources it
cites~\citep{gao2023alce,malaviya2024expertqa,niu2024ragtruth}. Their answer tasks do not expose an
upstream copying lineage whose dependence separates corroboration from echo
(Table~\ref{tab:compare}, Appendix~\ref{app:novelty}). Concurrent work reports that LLMs find repeated
evidence more persuasive than distinct support~\citep{naphade2026groupqa}. We study the complementary
question of whether the retrieved-answering step conditions the strength of its final claim on
upstream source dependence.
Measuring this cleanly is nontrivial: real claims are entangled with parametric knowledge, an
intuitive source-type definition can produce spurious failures
(Appendix~\ref{app:definitions}), and blanket caution can be
mistaken for discrimination. We address these threats with prior-free fictional claims, an audited source-type
definition, and matched \texttt{solid} controls.
Because several strong and inexpensive models resist the failure, the deployment question is where
the provenance decision should be made and how its conclusion should reach downstream agents. The experiments follow this decision path as three research questions. \textbf{RQ1} asks whether models can distinguish first-hand
from relayed evidence. \textbf{RQ2} asks whether ordinary answers use this distinction to calibrate
assertion strength. \textbf{RQ3} asks whether an explicit provenance conclusion can safely govern
cheaper downstream generation.

\paragraph{Contributions.}
\begin{enumerate}
\item \textbf{Controlled diagnostic.} We specify a task-specific policy for
  provenance-conditioned assertion and release \textbf{LaunderBench}: source-typing probes, 100
  model-gated prior-free weak-origin scenarios, \texttt{flagged} explicit-warning controls, and 96 matched
  \texttt{solid} independent-source controls.
\item \textbf{Behavioral findings.} In ordinary answers, the twelve tested models split into two
  cohorts: seven launder 0.29--0.96 of the weak-origin scenarios, and five launder 0.00--0.05. This
  cohort split holds under prompt, gate-family, and framework sensitivity checks, and it replicates
  for every commercial model when other model families write the scenarios. Every tested model reacts to the origin's own text, but only three
  shift decisively on the citation edges. Models can also identify provenance substantially better
  than they use it when answering.
\item \textbf{Explicit provenance handoff.} Where a model makes the right judgment, that judgment is lost in the handoff, not in the model.
  Asked directly, GPT-4o marks 97 of 100 weak-origin cases correctly, but when the next step receives
  only its written summary, 76 of 100 answers call the claim established, and attaching its own
  label cuts the 76 to 3. With a correct upstream provenance conclusion, a typed record yields
  0/400 weak-origin and 384/384 matched-solid establishments across four low-cost carriers.
\end{enumerate}

\section{Related Work}
\label{sec:related}

\paragraph{Theoretical foundations.}
Grammatical evidentiality
~\citep{aikhenvald2004evidentiality} shows that many languages obligatorily mark \emph{how} a speaker
knows what they assert, with reported/quotative values that parallel our first-hand/relayed split. Marking a source, however, is not the same as licensing an assertion. Norms of assertion
~\citep{williamson2000knowledge,lackey2007norms} and the epistemology of testimony
~\citep{coady1992testimony,lackey2008learning} motivate attention to warrant and transmission.
Sch\"utz's social distribution of knowledge~\citep{schutz1946wellinformed,schutz1973structures} and
rumor theory~\citep{allport1947rumor} describe how source type degrades with social transmission, a
dimension classical computational rumor models~\citep{daley1965stochastic} omit by treating the rumor
as an atomic token. Cognitive accounts of \emph{epistemic
vigilance} hold that widely shared belief adds evidential weight only when its holders arrived at it
independently, naming the acceptance of an unsourced but widely repeated claim an ``obvious
circularity''~\citep{sperber2010vigilance}. Expert fact-checkers, in turn, judge credibility by
\emph{lateral reading}, leaving a page to trace repeated claims to a common origin rather than counting
on-page corroboration~\citep{wineburg2019lateral}. The \emph{illusory-truth} effect shows humans,
too, mistake mere repetition for validity~\citep{hasher1977frequency,fazio2015knowledge}. These
literatures describe or prescribe, but none measures provenance licensing as a behavioral property of generative agents.

\paragraph{Attribution, provenance, and dependent sources.}
ALCE~\citep{gao2023alce} and ExpertQA~\citep{malaviya2024expertqa} evaluate cited long-form answers.
AIS formalizes whether text is \emph{attributable to identified sources}~\citep{rashkin2023ais}, and
AttributionBench and CiteEval study automatic, fine-grained citation assessment
~\citep{li2024attributionbench,xu2025citeeval}. RAGTruth instead annotates word-level hallucinations
relative to retrieved context~\citep{niu2024ragtruth}. These resources assess support by the cited or
retrieved context, not whether nominally distinct sources inherit one upstream origin.
Natural-language provenance work formalizes a claim's propagation graph and applies it to claim
verification~\citep{zhang2020provenance}, the text analogue of data-lineage standards for
records~\citep{w3c2013prov}. TROVE traces target sentences to source sentences and
classifies their transformation relationships~\citep{zhu2025trove}. Earlier database truth-discovery
work shows that copied values invalidate majority voting and explicitly models source dependence
~\citep{dong2009dependence}. These systems infer or evaluate provenance. We test whether a generator
converts visible dependence into a licensed assertion.

\paragraph{Evidence credibility and knowledge conflict.}
\citet{wan2024evidence} find that LLMs weight retrieved evidence by relevance-like surface features
rather than credibility. Knowledge-conflict studies show models readily adopt coherent-looking
external evidence~\citep{xie2024adaptive,wu2024clasheval}. Corpus-pollution work shows generated or
adversarial misinformation degrades retrieval-augmented answering
~\citep{pan2023misinformation,zou2025poisonedrag}, and deployed generative search engines frequently
mis-cite~\citep{liu2023verifiability}. These lines vary evidence content, quantity, or adversariality
rather than test whether an answerer discounts known dependence. Reliability-aware RAG estimates
source trustworthiness from agreement and aggregates by weighted vote~\citep{hwang2025rarag}. Absent dependence modeling, such agreement can overcount copied evidence~\citep{dong2009dependence}.
Self-RAG's reflection tokens~\citep{asai2024selfrag} grade per-passage support rather than inter-source
independence, as does self-verification prompting~\citep{dhuliawala2024cove}. Linguistic calibration matches
stated confidence to expected accuracy~\citep{mielke2022calibration,band2024calibration}. Our license instead conditions assertion strength on source lineage. GroupQA finds that
paraphrased repetition can outweigh distinct support~\citep{naphade2026groupqa}, but does not expose an
upstream citation chain. Wikipedia verifiability work taxonomizes why sentences require citations
~\citep{redi2019citation}, not whether several citations share a root. Concurrent FORCEBENCH uses
``citation laundering'' for a different failure, where a relevant passage under-warrants an
over-strong claim~\citep{qian2026forcebench}. We use \emph{source-dependence citation laundering} for
apparent corroboration inherited from one weak origin. Concurrent SearchGEO evaluates endorsement
corruption under adversarial web manipulation~\citep{chen2026searchgeo}, while origin-bound memory
work studies laundering across write--retrieve--act pipelines~\citep{louck2026originbound}. Our controlled task isolates shared-source echoes at answer-licensing time. Concurrent audits score
deep-research citations for accessibility and faithfulness~\citep{onweller2026cited}, use
\emph{source-dependence} for answer variation across conflicting institutional
sources~\citep{li2026samesource}, or show latent source preferences steer generations independently
of content~\citep{khan2026agentstrust}. None tests whether nominally distinct citations inherit one
weak origin. Character-knowledge and theory-of-mind
benchmarks, whose agent worlds resemble our provenance-tracing chains, bound temporal access or
reason about third-person beliefs
~\citep{ahn2024timechara,tu2024charactereval,kim2023fantom,he2023hitom,xu2024opentom,gu2024simpletom}. They do not constrain an answerer's own assertion by its evidence lineage.

\section{Task Definition}
\label{sec:formulation}

All tasks share one core: an agent holds evidence labeled with its source type and may assert only
what that evidence licenses. The answer policy is conservative and specific to this task: direct or
verified evidence may license assertion, a first-hand report licenses attribution, a relayed or unverified report licenses hedging, and inference must be marked as such. Source independence is
therefore one requirement among several. Sources that are independent but unreliable or unverified
still earn no assertion. This policy is an operational convention, instantiated entirely through
prompts and judges. Source types include
direct observation, testimony, hearsay, inference, environmental evidence, and role knowledge, among
others (Appendix~\ref{app:datasheet}).

\textbf{Testimony} means a report from a source
that directly established the fact, and \textbf{hearsay} means a report that passed through at least
one intermediary that did not itself establish the fact. ``X told me'' is locally compatible with
either case, so classification requires relating that report to its upstream lineage.
\textbf{Source-dependence citation laundering} is the violation of the answer policy in which a
multi-hop or circular chain is licensed as though its repeated mentions supplied independent
first-hand grounding.

\section{The LaunderBench Test Suite}
\label{sec:benchmark}

\paragraph{Why text, why synthetic.}
Text-based synthetic worlds give exact, per-claim ground-truth provenance that real systems cannot supply. Synthesis is therefore a deliberate choice, and four safeguards support its validity: naturalistic prose with
competing chains, checks that models have no prior knowledge of the claims, matched controls, and
stress tests across prompts and frameworks.

\subsection{Components}
\label{sec:components}

The core suite has two components. (a) \emph{Provenance-tracing chains}: a witness reports a fact to a second source, which passes it to the next, for up to six steps, and the
model must say how each source came to know the fact. There can be one chain or up to four competing chains that carry different answers. The text is either clean, with fixed template sentences, or naturalistic, with
varied wording. Each source's position in the chain fixes its correct source type: direct observation for the
witness, testimony for the second source, and hearsay for every later one.
(b) \emph{Citation-chain scenarios} (Figure~\ref{fig:design}): 100 retrieved-result bundles across 20
domains and three citation topologies (linear, star, fake corroboration) whose reliability is
assessed from both the origin's own text and the citation edges. These weak-origin bundles are the
\texttt{trace} condition. Component (b) also has two control conditions. In each \texttt{flagged} control, the
non-origin page that directly cites the root receives an explicit editor's note stating that the
figure is unverified and traces to one unsupported source, and 96 of the 100 scenarios have a
model-gated \texttt{solid} matched control.

\subsection{Model-gated scenario synthesis}
\label{sec:synthesis}

Scenario \emph{quality}, not quantity, is the gate. A strong author model (Claude Opus) drafts a
scenario for a chosen (domain, topology, origin-kind) cell. Citation edges are wired
programmatically to the requested topology (Appendix Figure~\ref{fig:synthesis_pipeline}). Three
checks admit a scenario.
\textbf{Programmatic checks:} no self-flagging ``tells'' (no page gives the setup away), no independent primary evidence on the non-origin pages, a different web domain for each page, a short and specific value stated in the claim, and no entity or value repeated across
scenarios. \textbf{Prior-freeness:} a
strong non-author validator (GPT-5.5, also on the evaluated panel) must report no prior knowledge of the entity or claim,
making the task a provenance test rather than a world-knowledge test. \textbf{Construct validity:} the same validator confirms the pages are non-doubting, the origin genuinely
weak, and every page traces to one origin. Each \texttt{solid} control is matched to an admitted
scenario in claim and domain, but replaces the weak lineage with ${\geq}2$ independent
primary sources under an analogous gate. All 100 target cells were filled, and 96 of the 100
scenarios have an admitted \texttt{solid} control (the other
four are excluded from \texttt{solid} denominators). The checks are selective: they rejected roughly 40\% of first
drafts. Blinded human annotators and a validator from another model family recheck the admitted
scenarios (Section~\ref{sec:exp_validity}, Appendices~\ref{app:tables} and~\ref{app:human}).

\subsection{Extending existing attribution benchmarks}
\label{sec:seeding}

Existing attribution datasets capture realistic question styles and evidence formats, but their claims
are often known to frontier models, confounding provenance reading with parametric knowledge. Our
adapter therefore extracts only a distributional \emph{seed} (a question-style exemplar, domain
hint, answer type, evidence format) from an ALCE, ExpertQA, or RAGTruth record. We use a seeded generator
to invent a new
prior-free fictional entity, value, and question under the same
\texttt{trace}/\texttt{flagged}/\texttt{solid}
controls. This extends attribution benchmarks along the source-graph dimension, from ``is the
answer supported?'' to ``are the supporting sources independent?'', without importing
world-knowledge contamination. Original examples are retained as \emph{native controls} under each
benchmark's own label semantics, reported separately.

\section{Metrics}
\label{sec:metrics}

We report four metrics: (i)~\emph{source-type accuracy} (by true type and chain depth), (ii)~its signature error, the \emph{over-attribution rate} (hearsay typed as testimony), (iii)~the \emph{laundering rate}, the proportion of distinct scenarios on which a poorly sourced
or circular claim is presented as established fact, and (iv)~for citation items, a
\emph{provenance-discrimination score},
$|\mathcal M|^{-1}\sum_{i\in\mathcal M}[I(\texttt{solid}_i{=}\text{established})-
I(\texttt{trace}_i{=}\text{established})]$, over the 96 scenarios that have both a \texttt{trace} and a \texttt{solid} version. This paired
contrast separates discrimination from blanket trust and blanket caution without mixing
denominators.

The two components are scored differently. On the provenance-tracing chains, the source types a
model assigns are checked against ground-truth labels. On the citation-chain scenarios, a judge model
reads the final answer and assigns one of four labels:
\emph{established}, \emph{hedged}, \emph{traced-weak-origin}, or \emph{absent}. The blinded audits
below and the real-web bank (Appendix~\ref{app:tables}) add a fifth label, \emph{debunked}, for
answers that give positive counter-evidence. Every label other than \emph{established} counts as
not established, so the primary metric is binary (established vs.\ not established). We audit
the production judge (GPT-4.1-mini) with stronger judges, a four-judge blinded panel, and a
blinded human audit (Appendix~\ref{app:human}). The stronger judges agree with it on
0.978--0.983 of binary decisions (Appendix~\ref{app:tables}).

\section{Experiments and Results}
\label{sec:experiments}

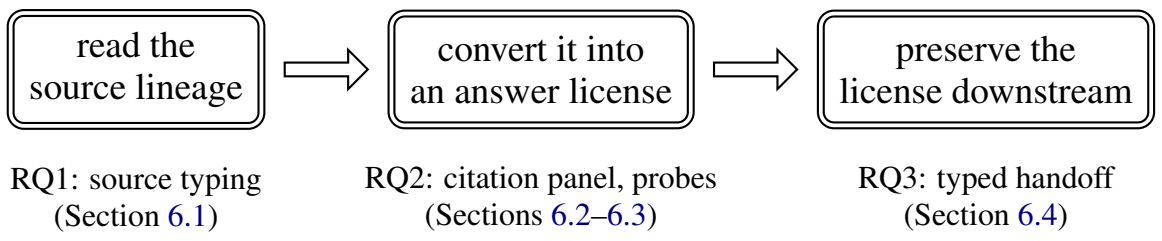
\begin{figure}[t]
\centering
\resizebox{\columnwidth}{!}{%
\begin{tikzpicture}[
  rstep/.style={draw, double, double distance=0.7pt, rounded corners=2.5pt, align=center,
               inner sep=3.5pt, font=\scriptsize, fill=white, minimum height=0.74cm},
  lab/.style={font=\tiny, align=center},
  blockarr/.style={single arrow, draw, fill=white, minimum height=0.55cm,
                   minimum width=0.26cm, single arrow head extend=1.3pt, inner sep=0.5pt}
]
\node[rstep] (a) {read the\\source lineage};
\node[blockarr, right=1.4mm of a] (ar1) {};
\node[rstep, right=1.4mm of ar1] (b) {convert it into\\an answer license};
\node[blockarr, right=1.4mm of b] (ar2) {};
\node[rstep, right=1.4mm of ar2] (c) {preserve the\\license downstream};
\node[lab, below=1.3mm of a] {RQ1: source typing\\(Section~\ref{sec:exp_typing})};
\node[lab, below=1.3mm of b] {RQ2: citation panel, probes\\(Sections~\ref{sec:exp_laundering}--\ref{sec:exp_mechanism})};
\node[lab, below=1.3mm of c] {RQ3: typed handoff\\(Section~\ref{sec:exp_handoff})};
\end{tikzpicture}}
\caption{The three decisions that ordinary answer accuracy conflates, and the experiment that
isolates each.}
\label{fig:roadmap}
\end{figure}

Ordinary answer accuracy conflates three distinct decisions: reading the source lineage, converting this lineage into an answer license, and preserving the license through downstream generation. We
separate these decisions in the order in which they occur
(Figure~\ref{fig:roadmap}). The model panel covers twelve models across four families,
including open-weight systems. Beyond the panel, we run additional reasoning models and off-the-shelf
frameworks to test alternative explanations for the cohort split (Section~\ref{sec:exp_validity}). Model identifiers, collection dates, and decoding settings are listed in
Appendix~\ref{app:repro}. Detailed statistical and per-category breakdowns are in
Appendix~\ref{app:tables}.

\subsection{Recovering provenance is possible but not general}
\label{sec:exp_typing}

In source typing, we ask a model to label each source in a chain with its source type. This
isolates reading the lineage from generating an answer. A model sees a witness $A_0$, a
first-hand report $A_0\rightarrow A_1$, and relayed reports $A_1\rightarrow A_2\ldots$, and it must
distinguish direct observation, testimony, and hearsay. The important error is not losing the value,
but promoting a relayed report to first-hand testimony, because this promotion grants the claim more
warrant than its lineage supplies.

\looseness=-1
We use two instruments to probe this. The first runs one uncontested chain to depth six over the whole panel,
rendered two ways: as shuffled sentences from which the model must work out who told whom, or as an
edge list that states the same chain outright. Nothing else differs.
Four models handle both renderings perfectly and Gemini Flash promotes a single relayed report,
so the task is solvable. The two open-weight models err only by calling first-hand testimony hearsay. The remaining
five promote relayed reports to first-hand testimony, and spelling the chain out for them helps
unevenly. GPT-4.1 and Haiku recover. GPT-4o-mini does not (41 of 60 relayed reports still called first-hand). GPT-4.1-mini and GPT-4o get worse, GPT-4o decaying from perfect at
every depth on prose to .25 by depth six once the edges are explicit
(Table~\ref{tab:cleanchain}). Being handed the structure is therefore not the same as reading it.

\looseness=-1
The second instrument asks whether clean-chain competence survives a messier setting: naturalistic
prose, a competing chain naming a different culprit, depth five. Three things move at once, so we use it
to grade difficulty rather than to isolate a cause. In this setup, Sonnet, Opus,
and o3 make no hearsay-to-testimony errors. GPT-4.1, Haiku, and GPT-4.1-mini over-attribute at
0.33--0.55. Small open models fail in the opposite direction, assigning nearly everything to
hearsay, a label-adherence failure rather than reliable typing.

\begin{findingbox}
\noindent\textbf{Finding 1 (provenance recovery is capability-gated).} Some models read source
lineage perfectly, while others promote relayed reports to first-hand testimony. This gradient
matters downstream: a model may be cheap and obedient without being qualified to make the
provenance decision itself.
\end{findingbox}

\subsection{Ordinary answers do not reliably use provenance}
\label{sec:exp_laundering}

On the citation-chain scenarios, we hold the value fixed and ask whether the lineage changes the
strength of the answer. We give each model five pages about a prior-free fictional claim, in an
order shuffled with a fixed seed. In \texttt{trace}, all
pages ultimately depend on one weak root. In \texttt{flagged}, a downstream page states this explicitly. In the matched \texttt{solid} condition, independent primary records support the same
claim. We count an \texttt{established} answer in \texttt{trace} as laundering.

\begin{table}[t]
\centering
\scriptsize
\setlength{\tabcolsep}{3pt}
\begin{tabular}{lccccc}
\toprule
\textbf{model} & \texttt{trace}$\downarrow$ & \textbf{95\% CI} & \texttt{flag.}$\downarrow$ & \texttt{solid}$\uparrow$ & \textbf{discr.} \\
\midrule
\textbf{GPT-4o-mini}  & \textbf{0.96} & [.90,\,.98] & 0.04 & 1.00 & $+$0.03 \\
\textbf{GPT-4.1-mini} & \textbf{0.74} & [.65,\,.82] & 0.01 & 1.00 & $+$0.24 \\
GPT-4o                & 0.72 & [.63,\,.80] & 0.01 & 1.00 & $+$0.27 \\
Qwen2.5-32B           & 0.70 & [.60,\,.78] & 0.01 & 1.00 & $+$0.28 \\
GPT-4.1               & 0.62 & [.52,\,.71] & 0.00 & 1.00 & $+$0.38 \\
Qwen3-30B             & 0.53 & [.43,\,.62] & 0.01 & 0.97 & $+$0.43 \\
Gemini-2.5-Flash      & 0.29 & [.21,\,.39] & 0.00 & 1.00 & $+$0.71 \\
o3                    & 0.05 & [.02,\,.11] & 0.00 & 1.00 & $+$0.95 \\
Claude Haiku 4.5      & 0.02 & [.01,\,.07] & 0.00 & 1.00 & $+$0.98 \\
GPT-5.5               & \textbf{0.00} & [0,\,.04] & 0.00 & 0.96 & $+$0.96 \\
Claude Sonnet 4.6     & \textbf{0.00} & [0,\,.04] & 0.00 & 0.79 & $+$0.79 \\
Claude Opus 4.8       & \textbf{0.00} & [0,\,.04] & 0.00 & 0.92 & $+$0.92 \\
\bottomrule
\end{tabular}
\caption{First screening result: source-dependence citation laundering on 100 prior-free
weak-origin scenarios
($n{=}100$ for \texttt{trace}/\texttt{flagged}, $n{=}96$ for \texttt{solid}). Discrimination is the
paired mean \texttt{solid}--\texttt{trace} difference on the 96 matched scenarios. Stronger judges
shift closed-model rows ${\le}.01$, and the Qwen rows are lower bounds (Appendix~\ref{app:tables}).
The rate is an outcome over the composite weak-origin condition. The probe in
Section~\ref{sec:exp_mechanism} separates its origin-text and edge channels. Rates are
instrument-relative (Table~\ref{tab:author}). GPT-5.5 is also the gate validator
(Section~\ref{sec:synthesis}).}
\label{tab:panel}
\end{table}

Table~\ref{tab:panel} is a first screening result. It separates three behaviors. GPT-4o-mini accepts both weak and solid bundles, so
its low discrimination reflects blanket trust. GPT-5.5, o3, and the Claude models reject the weak
lineage while accepting most solid controls. This is the behavior that our benchmark targets.
Sonnet adds the third behavior, cautious answering: it is more conservative on \texttt{solid}, but it still reports the
value and withholds only the established license. In 13 of the 96 \texttt{solid} scenarios, its
rejection cites the placeholder \texttt{.example} domains of the pages (Appendix~\ref{app:sandbox}),
and under a neutral prompt it accepts 94 of 96 (Appendix~\ref{app:tables}). Most importantly, \texttt{flagged} is near zero across the
panel, so the models can use a reliability conclusion once it is surfaced. They differ in whether they
recover and apply the same conclusion from the citation lineage.

The cohort split persists across topologies, domains, seed corpora, and three scenario-author
families. Linear chains are generally
hardest, but laundering appears in every domain, and the split recurs on panels seeded from
ExpertQA, ALCE, and RAGTruth: GPT-4o-mini launders 13--21 of 30 weak-origin items, whereas Opus
launders none and still accepts most solid controls.

\begin{findingbox}
\noindent\textbf{Finding 2 (the boundary is provenance use, not general caution).} The panel
does not form a simple accuracy ranking. It separates blanket trust, lineage-sensitive answering,
and cautious answering. The large gap between \texttt{trace} and \texttt{flagged} locates the main
failure in implicit use of the weak-origin evidence rather than an inability to follow an
explicit warning.
\end{findingbox}

\subsection{The bottleneck lies between diagnosis and generation}
\label{sec:exp_mechanism}

\begin{figure}[t]
\centering
\begin{tikzpicture}
\begin{axis}[
  width=\columnwidth, height=3.15cm,
  ymin=-1, ymax=30,
  ytick={0,10,20,30},
  ylabel={laundering / 30$\downarrow$},
  ylabel style={font=\scriptsize},
  xmin=-0.4, xmax=7.4,
  xtick={0,1,2,3,4,5,6,7},
  xticklabels={raw, trace, diag.+ans., self-JSON, graph, roots, type, license},
  x tick label style={rotate=34, anchor=east, font=\scriptsize},
  y tick label style={font=\scriptsize},
  legend style={font=\scriptsize, at={(0.5,1.03)}, anchor=south, legend columns=3,
    draw=none, fill=none, /tikz/every even column/.append style={column sep=5pt}},
  legend cell align={left},
  ymajorgrids, grid style={gray!20},
  tick align=outside,
  clip=false
]
\addplot+[color=blue!70!black, mark=*, mark size=1.6pt, semithick, solid] coordinates
  {(0,26) (1,12) (2,15) (3,4) (4,22) (5,6) (6,0) (7,0)};
\addlegendentry{4o-mini}
\addplot+[color=orange!90!black, mark=square*, mark size=1.5pt, semithick, dashed] coordinates
  {(0,21) (1,1) (2,0) (3,0) (4,9) (5,0) (6,0) (7,0)};
\addlegendentry{4.1-mini}
\addplot+[color=purple!75!black, mark=triangle*, mark size=1.8pt, semithick, densely dotted] coordinates
  {(0,1) (1,0) (2,0) (3,0) (4,0) (5,0) (6,0) (7,0)};
\addlegendentry{Haiku 4.5}
\end{axis}
\end{tikzpicture}
\caption{An explicit source type or answer license eliminates laundering for every probed model,
while a citation graph alone barely helps GPT-4o-mini. The ladder progressively exposes a tracing
instruction, same-context diagnosis, self-produced diagnostic JSON, graph, roots, type, or license
(30-scenario subset, while Table~\ref{tab:panel} uses all 100).}
\label{fig:mechanism}
\end{figure}
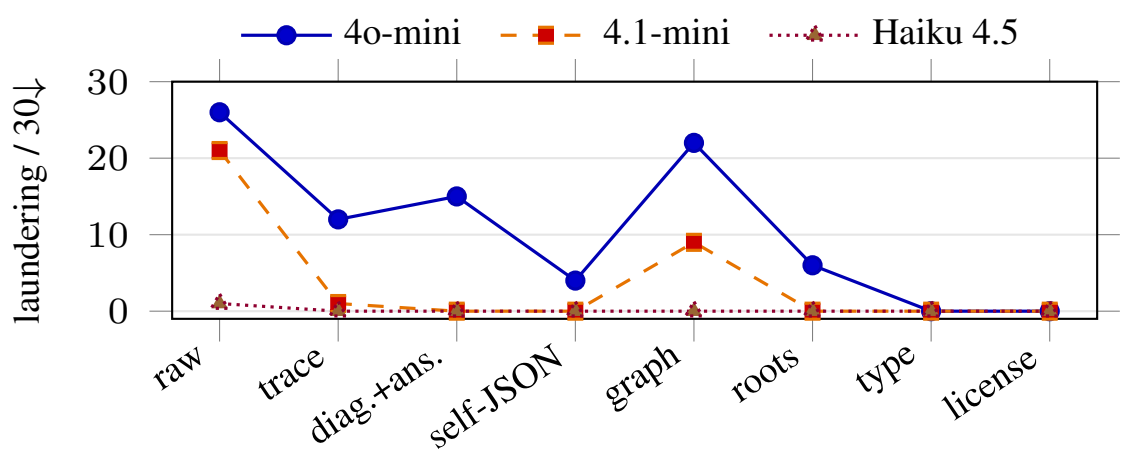

The citation panel leaves two explanations open: a model may fail to recover the graph, or may recover
it without letting this judgment control its answer. The intervention ladder in
Figure~\ref{fig:mechanism} separates them by exposing progressively more provenance state.
If visibility were the only problem, the graph and roots should close the gap. If answer-time use is
the problem, the distilled type or license should matter more.

All three probed models (GPT-4o-mini, GPT-4.1-mini, and Haiku) diagnose citation edges, roots, and
source type substantially better than their raw answers suggest (Appendix~\ref{app:tables}). GPT-4o-mini still launders 22/30 with the graph made explicit and 6/30 with the root
set, but falls to 4/30 when it carries its own diagnostic JSON and to 0/30 when given the source type
or license. GPT-4.1-mini needs only the directed tracing instruction, and Haiku is already near zero.
Thus the useful intervention differs by model, but generic access to more representation is not
enough for the heaviest launderer. The provenance conclusion must become answer-controlling state.

The explanations models give are consistent with this reading. Among the six heaviest launderers,
79\% of established answers explicitly appeal to agreement across multiple sources. They are not
merely omitting a caveat. They are treating multiplicity as independence.

\begin{table}[t]
\centering
\scriptsize
\setlength{\tabcolsep}{1.8pt}
\begin{tabular}{lcccc}
\toprule
\textbf{model} & \makecell{\textbf{dep.}\\\textbf{estab.}} & \makecell{\textbf{ind.}\\\textbf{estab.}} & \makecell{\textbf{paired diff.}\\\textbf{[95\% CI]}} & \makecell{\textbf{root diag.}\\\textbf{dep.; ind.}} \\
\midrule
GPT-5.5           & 4  & 93  & \textbf{+.89} [.83, .95] & 100; 100 \\
o3                & 12 & 71  & \textbf{+.59} [.49, .68] & 100; 100 \\
Claude Opus 4.8   & 0  & 50  & \textbf{+.50} [.40, .60] & 88; 100 \\
\midrule
GPT-4.1           & 89 & 100 & +.11 [.05, .17]    & 91; 99 \\
Claude Sonnet 4.6 & 53 & 62  & +.09 [.01, .17]    & 69; 100 \\
Claude Haiku 4.5  & 68 & 75  & +.07 [.00, .14]    & 17; 88 \\
Gemini 2.5 Flash  & 98 & 98  & +.00 [$-$.03, .03] & 84; 97 \\
GPT-4o            & 100 & 100 & +.00 [.00, .00]   & 59; 100 \\
GPT-4.1-mini      & 100 & 100 & +.00 [.00, .00]   & 35; 92 \\
GPT-4o-mini       & 99 & 99  & +.00 [.00, .00]    & 0; 0 \\
Qwen2.5-32B       & 72 & 66  & $-$.06 [$-$.12, .00] & 4; 37 \\
Qwen3-30B         & 76 & 69  & $-$.07 [$-$.13, $-$.01] & 1; 36 \\
\bottomrule
\end{tabular}
\caption{Graph-only rewiring probe. Among the twelve panel models, only GPT-5.5, o3, and Opus shift
decisively on the citation edges. Each of the 100 pairs has two arms: in the dependent arm (dep.)
the pages share one origin, and in the independent arm (ind.) they rest on two independent origins.
All counts are out of 100 scenarios per arm, and rows are ordered by the paired difference
(independent minus dependent). Established counts
use the production judge. Paired CIs bootstrap scenarios (10,000 resamples). The root diagnostic
jointly requires unique, exact IDs and the correct declared count. Both answer judges are blind to
arm and expected graph. Appendix~\ref{app:tables} describes the second judge and the root diagnostic.}
\label{tab:graphonly}
\end{table}

Whether near-zero laundering reflects edge use still needs its own test, and this test gives the
main result of this section (Table~\ref{tab:graphonly}). A matched graph-only
rewiring probe (100 pairs, every page written in the same neutral way, with the two arms of a pair differing in exactly
two citation pointers) separates the answer policies of all twelve panel models directly.
GPT-5.5, o3, and
Opus shift their answer license on the edges alone ($+.89$, $+.59$,
$+.50$, with exact McNemar $p\le1.8\times10^{-15}$). No other model flips half its pairs: four
establish both arms at ceiling, Haiku, Sonnet, and GPT-4.1 still establish 53--89\% of dependent
bundles ($+.07$--$+.11$, with GPT-4.1's and Sonnet's reliable), and the two open-weight models drift the other way without reaching significance
($-.06$, $-.07$, with $p\ge.065$).

A factorial rerun crosses the same pairs with the origin's own text and completes
the split: once the root pages call themselves unverified guesses, every model abandons the
independent bundles it had accepted and none establishes more than 4 of 100 in either arm. Decisive edge shifts belong to three panel models and a Google frontier
probe ($+.81$, Appendix~\ref{app:tables}). Low laundering on the citation panel therefore certifies decisive
edge use only for GPT-5.5, o3, and Opus. For Haiku and Sonnet it reflects restraint keyed to the origin's own
text.

\begin{findingbox}
\noindent\textbf{Finding 3 (diagnosis can be better than answer-time use).} For the models that
launder, a recoverable provenance judgment does not automatically constrain generation. Directed
provenance state succeeds where same-context diagnosis or a bare graph does not, making the
diagnosis-to-generation boundary the relevant intervention point.
\end{findingbox}

\subsection{Explicit provenance carries the decision downstream}
\label{sec:exp_handoff}

If the bottleneck is using the provenance decision, the system need not ask every downstream model
to rediscover it. We separate \emph{typing}, which reads the pages and assigns a provenance state,
from \emph{carrying}, which maps an already-typed state to the answer license. At answer time the carrier does not see
the pages: it receives its own prose briefing, its own briefing plus its own typed field, or a
briefing and record produced once by a qualified typer. Withholding the pages is the deployment
shape under test: one component reads the sources once, cheaper ones consume its output. The same
intervention with the pages present is the ladder's \emph{type} arm above.

\begin{table}[t]
\centering
\scriptsize
\setlength{\tabcolsep}{3.2pt}
\begin{tabular}{lcccc}
\toprule
\textbf{carrier} & \makecell{\textbf{own field}\\\textbf{weak/circ.}} & \makecell{\textbf{self}\\\textbf{downstr.}$\downarrow$} & \makecell{\textbf{self}\\\textbf{carry}$\downarrow$} & \makecell{\textbf{Opus typed}\\\textbf{handoff}$\downarrow$} \\
\midrule
GPT-4o-mini  & 23/100 & 96/100 & 73/100 & 0/100 \\
GPT-4.1-mini & 55/100 & 76/100 & 44/100 & 0/100 \\
GPT-4o       & \textbf{97/100} & \textbf{76/100} & \textbf{3/100}  & 0/100 \\
GPT-4.1      & 92/100 & 78/100 & 8/100  & 0/100 \\
\bottomrule
\end{tabular}
\caption{Where a model makes the right judgment, that judgment is lost in the handoff, not in the model. Asked directly, GPT-4o detects 97 of
100 weak or circular scenarios (\emph{own field}). With only its own prose briefing, 76 of 100
answers launder (\emph{self downstr.}), and with its own typed field added, 3 do (\emph{self
carry}). With the qualified typer's record (Opus flags all 100 weak-origin bundles), every carrier
avoids establishment, and all four establish all 96 matched solids.}
\label{tab:handoff}
\end{table}

The three arms locate the gain. Default prose leaves laundering at 0.76--0.96.
Adding the carrier's own field helps only to the extent that the carrier typed the source correctly.
Replacing this field with the qualified typer's decision gives 0/100 weak-origin establishments
per carrier and 384/384 solid establishments, which is perfect paired discrimination
throughout. These zeros are instruction following: the answering model simply follows the label it
is given. A gpt-5-mini typer reproduces
the four zeros, so the result is not tied to Opus.

We use a deliberately weak typer to check that the field itself drives the carriers' answers. This typer misclassifies 41 weak-origin scenarios as independent, after which the carriers establish 39--41 of the same items. Across these runs, carriers establish 159/164 cases when the field is wrong and 0/236 when it is correct, so the field controls downstream behavior. We also test whether the format of the record matters.
Instructed prose, a briefing required to state how the value is grounded, also scores
0/400 on weak origins but establishes only 344/384 solid controls, as generated prose adds
qualifications beyond the typing decision. Writing the same record out as a fixed
English sentence instead of JSON changes nothing (0/400 and 384/384), so what the record buys is a
decision downstream generation cannot rewrite, not a particular encoding.

Two cases fall outside these experiments. First, the label has no option for one real source that
other pages repeat, so we add the number of independent sources to the released record. Second, each
handoff here carries a single claim, and we did not test multi-claim reports.

\begin{findingbox}
\noindent\textbf{Finding 4 (type once, carry the license).} Conditional on a correct provenance
decision, low-cost carriers preserve the licensed behavior in both directions. The remaining system
problem is therefore choosing and auditing the typer, not asking each carrier to re-infer lineage
from prose.
\end{findingbox}

\subsection{Alternative explanations do not account for the split}
\label{sec:exp_validity}
\label{sec:exp_metric_validity}

Benchmark construction, the judge, reasoning effort, and framework scaffolding do not account for
the cohort split. Appendix~\ref{app:tables} reports these checks.

Together these controls support a specific behavioral account. The citation lineage is
available, many models can diagnose it, and all models can follow its conclusion once made
explicit under the panel prompt. The boundary is whether ordinary answering turns this lineage into appropriate assertion strength.

\section{Implications}
\label{sec:discussion}

\paragraph{Provenance is a control variable, not just metadata.}
Provenance text in the context is not enough. What matters is the answer license
derived from it. This changes the unit of evidence, because
retrieval presents documents, but documents need not be independent
observations. Five pages that inherit one unsupported report add visibility, not warrant. Two primary records that do not inherit from one another can
change the license even with an identical surface claim. Provenance-aware
answering must collapse apparent support by lineage before deciding how
strongly to speak. A support-only attribution score can miss this, because every
citation may entail the answer while all the cited pages trace back to one weak
origin.

\paragraph{The system boundary can move.}
With the license explicit, the downstream task is simpler and auditable.
Structured records are not intrinsically more truthful than prose. Their
advantage is operational. The field is obligatory, machine-checkable,
conflict-robust (Appendix~\ref{app:tables}), and symmetric between
withholding weak claims and establishing supported ones. The operative property is that the decision is carried unaltered, not that it is JSON
(Section~\ref{sec:exp_handoff}). For single-use
queries, direct answering by a qualified typer may remain simpler.

The handoff is a contract. When the upstream typer supplies the wrong
license, carriers faithfully propagate it, and with the right one, their
confident prose no longer overrides it. Reliability has moved upstream rather
than appeared for free. This relocation is useful because a system can qualify
one component on weak and solid inputs, log its per-claim decision, and reuse
it across consumers.

\paragraph{The benchmark is a decomposition diagnostic.}
LaunderBench is a behavioral decomposition (recover lineage, derive a
license, carry the license), not a leaderboard of research ability. The
matched contrast asks whether assertion strength changes with the evidential
basis, and the typing and intervention tasks explain why.

\section{Conclusion}
\label{sec:conclusion}

\looseness=-1
Every probed model reacts to the origin's own text, but only three shift decisively on the citation
edges. Repeated citations are not independent evidence. LaunderBench makes this distinction a controlled
task and separates recovering a lineage from using it to govern an answer. Seven of the twelve tested
models launder such echoes in ordinary answers. An explicit license
resolves the gap. With correct provenance, carriers score 0/400 weak-origin and 384/384 solid.
Systems should preserve not just what sources say but the provenance decision that sets what
they license. As research pipelines add hops between evidence and answer, every relay is another
chance to mistake an echo for corroboration.\label{end:main}

\section*{Limitations}

\paragraph{Construct and instrument.}
The main \texttt{trace}--\texttt{solid} answer-license contrast jointly changes upstream dependence,
root warrant, terminal count (pages that cite nothing), and some page text, so it is not an
edge-only causal estimate. The
matched graph-only probe isolates rewiring across all twelve panel models, and its origin-rewriting
factorial separates the origin-text and edge channels, but both use one template family with one
generation per cell. Models that establish or withhold in both arms at a given level of origin text may hold licensing
thresholds that level cannot cross, so their edge insensitivity is an operating-point observation,
not proof they never consult edges.

\paragraph{Synthetic and temporal scope.}
Fictional English scenarios provide exact ground truth and limit parametric knowledge, but they do not
represent the full ambiguity, source reputation, or multilingual evidential marking of the live web.
The real-web bank in Appendix~\ref{app:tables} is small (twenty documented incidents), manually
curated, and not prior-free. Our data therefore cannot say how common citation laundering is on the open web. Model
results describe the deployed models and provider configurations collected on the dates in
Appendix~\ref{app:repro}. They do not establish a persistent tier, laboratory, or training-recipe
law. Across three scenario-author families, cohort membership replicates for every commercial
model, but absolute rates and discrimination margins are properties of the item population, and
the strictest models discount one author family's \texttt{solid} controls, an effect that is
mostly prompt-conditional (Appendix~\ref{app:tables}).

\paragraph{Missing citations.}
Every LaunderBench bundle shows which page cites which, and we did not test missing citations. A
deployed system must first find which page copied which. Explicit links and quotations show this
directly, publication dates give the direction, and similar text indicates copying when no link
remains. Each signal can be absent or wrong, so the recovered graph is itself an estimate.

\paragraph{Evaluation uncertainty.}
The fixed production judge is strongly audited, but fine-grained non-laundering labels retain
annotator variation.
Small open-model source typing is additionally confounded by format adherence. Most commercial runs record one answer per scenario under fixed or provider-default decoding, so scenario-level intervals
do not estimate uncertainty over all generation seeds or future model aliases; a three-repetition
probe on the five strictest models reproduces every near-zero \texttt{trace} cell and tracks
\texttt{solid} within three items (Appendix~\ref{app:tables}).

\paragraph{Framework scope.}
The framework replications inject all pages without realistic retrieval, ranking, or truncation, and the research-agent experiment
(Appendix~\ref{app:tables}) uses a controlled mock-search endpoint. They are prompt and planning
stress tests, not end-to-end evidence about open-web research behavior.

\paragraph{Handoff scope.}
Instructed prose matches typed-record safety on weak origins but yields 344/384 rather than 384/384
matched-solid establishments. The record's demonstrated role is a stable, machine-checkable
bidirectional license plus conflict handling (Appendix~\ref{app:tables}).
Production integration, multi-claim records, adversarial typing, and schema enforcement remain
untested. The licensing policy is a
conservative task convention: provenance independence is necessary but not sufficient for truth.

\paragraph{Negative results.}
Four internal hypotheses failed clean controls (documented in Appendix~\ref{app:tables}); those
failures delimit the claim to source-dependence use in ordinary answering and the explicit state
that repairs it, not any controller, prompt, or model-family advantage.

\section*{Ethical Considerations}
\label{sec:ethics}

\paragraph{Human annotation.}
The blinded human audits (the 112-item citation-answer audit and the 40-scenario construct audit;
Appendix~\ref{app:human}) were completed by five adult volunteer annotators, four of whom had not
been exposed to the paper's hypotheses before annotating. Every annotator gave verbal informed
consent covering research use and release of the anonymized labels before starting; the annotation
interface additionally requires acknowledging the task rules (no AI assistance, no web search)
before any item is shown, and hides model, condition, panel, and machine-judge labels from every
annotator. Annotator-exclusion robustness analyses, including exclusion of the one
hypothesis-aware annotator (the study designer), appear in Appendix~\ref{app:human}.
Annotation was uncompensated volunteer work by adult colleagues, none with a reporting-line
relationship to the study designer; annotators could pause or stop at any point, and recorded
on-item time for the full 112-item pass had a median of roughly 2.7 hours per annotator
(per-item telemetry; the 40-scenario construct audit was not time-instrumented). Annotators labeled only machine-generated answers and synthetic scenario pages about
fictional entities or documented public circular-reporting incidents (the 40 real-web case-bank
answer records); no personal, demographic, or otherwise sensitive data were collected, and no vulnerable
populations were involved. We assess the activity as minimal-risk annotation of machine output, and
on that basis did not seek an institutional review-board determination.

\paragraph{Artifacts and licenses.}
ALCE, ExpertQA, and RAGTruth are used under their MIT licenses, consistent with their intended
research use. The seeded panels are generated from distributional seeds (question style, domain
hint, answer type, evidence format): the generation prompt quotes the source question only as a
style exemplar, and the model-visible content of every released seeded scenario shares no
sentence-level text with its source record (verified programmatically;
Appendix~\ref{app:tables}). Seed-provenance
metadata retains the quoted source question and short evidence-format snippets under the same MIT
licenses; native-control items
reproduce original examples under the same licenses. All LaunderBench scenarios concern invented
entities with placeholder (\texttt{.example}) domains, so the benchmark neither asserts anything
about real people or organizations nor releases plausible real-world misinformation. Commercial
models were accessed through their official APIs under the providers' terms of service; open-weight
models were run locally.

\paragraph{Dual use.}
The failure we measure (circular sourcing accepted as corroboration) could in principle inform
adversarial content placement (e.g., data-void
attacks~\citep{golebiewski2019datavoid} on research agents). We judge disclosure
net-positive: the failure is already exploitable by anyone who can publish web pages, whereas the
paper contributes controlled measurement and two mitigations (surfacing reliability flags and
type-once-carry handoff) that defenders can deploy.

\bibliography{refs}

\appendix

\section{Datasheet (Summary) and License}
\label{app:datasheet}

This summary follows \emph{Datasheets for Datasets}~\citep{gebru2021datasheets}; the full
datasheet (version 2026-07-10) ships with the repository.
\textbf{Purpose:} evaluate whether an LLM agent asserts only what its source type permits, and measure
omniscience leakage (asserting a proposition absent from the agent's own evidence), false-witness,
and rumor-as-fact in partially observable multi-character worlds.
\textbf{Composition (ecological set):} 250 LLM-authored worlds; 1{,}575 queries; 5--7 agents per world
(avg.\ 6.3); 126 distinct answer atoms; 1{,}242 provenance-typed claims (direct\_observation .25,
testimony .21, hearsay .20, environmental\_evidence .17, role\_knowledge .16, other .02); distorted
rumor in 248/250 worlds; 21\% no-knowledge agents that must abstain. Every claim carries (proposition,
source\_type, source\_agent, time, reliability, verification, value); every query carries acceptable
modes, forbidden claims, and the true value. \textbf{Collection:} worlds authored by GPT-4.1-mini under
a strict programmatic validator (265 attempts $\rightarrow$ 250 accepted); single-author-family
provenance is a stated limitation. \textbf{Sensitive content:} none (fictional characters; no real
persons or PII). \textbf{Citation-chain items} are authored by Claude-Opus and
gated by GPT-5.5 (prior-freeness and construct validity), with topologies wired in code;
Table~\ref{tab:composition} gives the canonical-100 domain and topology composition.
\textbf{Diversity (canonical-100):} structure is fixed by design (five pages per condition;
topologies wired in code) while surface realization varies: all 100 entities and claimed values are
unique, origin pages instantiate five distinct weak-origin kinds, and scenario corpora share little
text (mean pairwise content-word Jaccard 0.077 \texttt{trace} / 0.077 \texttt{solid}, bigram
0.008/0.003, maxima 0.31/0.27; corpus-wide distinct-trigram ratio 0.92 in both splits). Exact-sentence
reuse ($\geq$6 words) across the 100 \texttt{trace} corpora is a single two-scenario forum-idiom line
(likewise one boilerplate line among \texttt{solid} corpora); statistics:
\texttt{experiments/scenario\_diversity\_stats.py}.
\textbf{Release:} generators, the world set, the 100 canonical + 3$\times$30 seeded citation scenario
sets with matched \texttt{solid} controls, all retained row-level outputs plus the documented
legacy aggregates (Appendix~\ref{app:repro}), and this datasheet; intended release licenses
are CC-BY-4.0 for data and MIT for code, subject to final repository packaging.
\textbf{Maintenance:} scenario sets are versioned and frozen per paper revision (v2 here);
corrections and extensions go through the release repository's issue tracker, and the author/gate
pipeline can regenerate any cell suspected of contamination. The closed-book probe
(Appendix~\ref{app:tables}) is the release's standing contamination check: a post-release model that
recovers canonical values closed-book above chance triggers a regenerated, re-gated version bump.

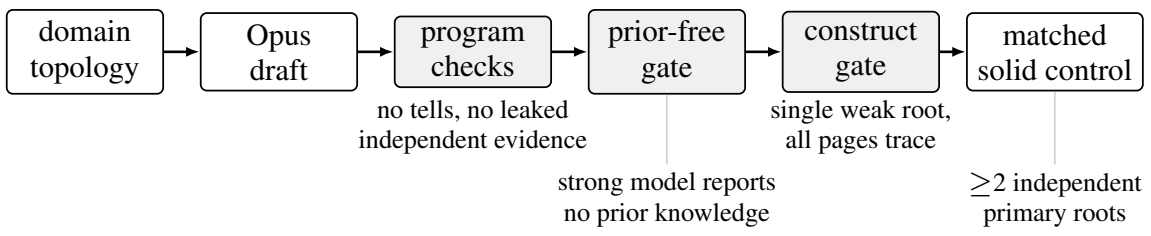
\begin{figure}[t]
\centering
\resizebox{\columnwidth}{!}{%
\begin{tikzpicture}[
  stage/.style={draw, rounded corners=1.4pt, align=center, inner sep=2.4pt,
                minimum width=1.25cm, minimum height=0.54cm, font=\scriptsize},
  gate/.style={stage, fill=gray!12},
  arr/.style={-{Latex[length=1.1mm]}, semithick}
]
\node[stage] (seed) at (0,0) {domain\\topology};
\node[stage] (draft) at (1.55,0) {Opus\\draft};
\node[gate]  (prog) at (3.10,0) {program\\checks};
\node[gate]  (prior) at (4.65,0) {prior-free\\gate};
\node[gate]  (construct) at (6.20,0) {construct\\gate};
\node[stage] (solid) at (7.75,0) {matched\\solid control};
\draw[arr] (seed) -- (draft);
\draw[arr] (draft) -- (prog);
\draw[arr] (prog) -- (prior);
\draw[arr] (prior) -- (construct);
\draw[arr] (construct) -- (solid);
\node[font=\tiny, align=center] at (3.10,-0.60) {no tells, no leaked\\independent evidence};
\node[font=\tiny, align=center] at (4.65,-1.18) {strong model reports\\no prior knowledge};
\node[font=\tiny, align=center] at (6.20,-0.60) {single weak root,\\all pages trace};
\node[font=\tiny, align=center] at (7.75,-1.18) {$\geq$2 independent\\primary roots};
\draw[gray!60, very thin] (prior.south) -- (4.65,-0.90);
\draw[gray!60, very thin] (solid.south) -- (7.75,-0.90);
\end{tikzpicture}
}
\caption{Canonical scenario construction pipeline. Topologies are wired in code, while independent
quality gates filter drafts for leakage, prior knowledge, and construct validity before a matched
\texttt{solid} control is admitted.}
\label{fig:synthesis_pipeline}
\end{figure}

\begin{table}[t]
\centering
\scriptsize
\setlength{\tabcolsep}{3pt}
\begin{tabular}{lcccc}
\toprule
\textbf{domain (abridged)} & \textbf{lin.} & \textbf{star} & \textbf{fake} & $n$ \\
\midrule
animal biology (lifespan, speed, diet) & 1 & 2 & 2 & 5 \\
demographic / economic statistic & 2 & 2 & 1 & 5 \\
geographical superlative & 2 & 2 & 1 & 5 \\
physical/chemical property of a material & 1 & 2 & 1 & 4 \\
attribution of a famous saying & 2 & 1 & 2 & 5 \\
regional rainfall / record temperature & 2 & 1 & 2 & 5 \\
birthplace of a historical figure & 2 & 2 & 2 & 6 \\
composer / premiere year of a piece & 1 & 2 & 2 & 5 \\
rule / scoring of a traditional game & 2 & 2 & 1 & 5 \\
founding year of a small town & 2 & 2 & 2 & 6 \\
height of a monument / tower / statue & 1 & 2 & 2 & 5 \\
credit for a theorem / constant / law & 1 & 2 & 2 & 5 \\
meaning of a flag / crest symbol & 2 & 2 & 1 & 5 \\
moons / rings / distance of a body & 2 & 1 & 2 & 5 \\
etymology of a word or phrase & 2 & 1 & 2 & 5 \\
record time / margin in a competition & 2 & 2 & 1 & 5 \\
ingredient / origin of a regional dish & 2 & 1 & 2 & 5 \\
year of a battle / treaty / coronation & 2 & 1 & 1 & 4 \\
invention year of a tool / product & 2 & 1 & 2 & 5 \\
first person to achieve a feat & 1 & 2 & 2 & 5 \\
\midrule
total & 34 & 33 & 33 & 100 \\
\bottomrule
\end{tabular}
\caption{Canonical-100 composition: 20 domains $\times$ 3 topologies (linear / star / fake
corroboration), 4--6 scenarios per domain. Domain cells are abridged; the verbatim domain prompts
ship with the scenario files.}
\label{tab:composition}
\end{table} The model-visible content of the seeded panels
shares no sentence-level text with the ALCE / ExpertQA / RAGTruth source records (all MIT-licensed);
seed-provenance metadata retains the quoted source question under the same MIT licenses, and original
examples from those datasets appear only in the native-control evaluations.

\section{Positioning Audit (Condensed)}
\label{app:novelty}

\begin{table}[t]
\centering
\scriptsize
\setlength{\tabcolsep}{3pt}
\begin{tabular}{@{}>{\raggedright\arraybackslash}p{3.1cm}ccc@{}}
\toprule
\textbf{research line} & \makecell{\textbf{prior-free}\\\textbf{control}} & \makecell{\textbf{upstream}\\\textbf{dependence}} & \makecell{\textbf{final-answer}\\\textbf{license}} \\
\midrule
Claim provenance graphs~\citep{zhang2020provenance} & $\times$ & $\checkmark$ & $\times$ \\
Copy-aware truth discovery~\citep{dong2009dependence} & $\times$ & $\checkmark$ & $\times$ \\
Attribution / provenance eval.~\citep{xu2025citeeval,zhu2025trove} & $\times$ & $\checkmark$ & $\times$ \\
Repetition persuasion~\citep{naphade2026groupqa} & $\times$ & partial & $\checkmark$ \\
Warrant laundering~\citep{qian2026forcebench} & $\times$ & $\times$ & $\checkmark$ \\
\textbf{LaunderBench (ours)} & $\checkmark$ & $\checkmark$ & $\checkmark$ \\
\bottomrule
\end{tabular}
\caption{Closest research lines and LaunderBench's intersection. Earlier work represents claim
provenance, models copying dependence, or evaluates citation warrant; LaunderBench combines a
prior-free controlled task, upstream dependence, and the license expressed in a generated answer.
``Partial'' denotes repeated-vs.\ distinct evidence without an explicit citation lineage.}
\label{tab:compare}
\end{table}

The closest technical antecedents are natural-language claim provenance graphs
~\citep{zhang2020provenance}, copy-aware truth discovery~\citep{dong2009dependence}, and attribution
evaluations that model provenance, credibility, or redundancy~\citep{xu2025citeeval,zhu2025trove}.
They establish that source lineage and dependence matter; our narrower contribution is to measure
whether a generative answerer uses an available lineage when choosing assertion strength. GroupQA
varies repeated versus distinct support~\citep{naphade2026groupqa}; LaunderBench adds explicit
upstream citation dependence and prior-free claims. FORCEBENCH independently uses ``citation
laundering'' for a passage--claim warrant mismatch~\citep{qian2026forcebench}; our term denotes
\emph{source-dependence citation laundering}, in which several apparent sources share a weak origin.
The current \texttt{trace}/\texttt{solid} contrast changes both lineage and root warrant, so it tests
their composite use rather than an edge-only graph intervention.

\section{Source-Type Definitions and the Gold Audit}
\label{app:definitions}

The following labels are operational categories for the controlled task. In particular,
\texttt{testimony} means a first-hand report under the prompt below; it does not assert that
philosophical testimony generally requires direct observation or never transmits knowledge.

\paragraph{Corrected definition (verbatim from the typing prompt).}
\begin{quote}\small\ttfamily
source\_type is one of: direct\_observation = they personally witnessed it; testimony = they were told
by the person who personally witnessed it (a first-hand source); hearsay = the news reached them
through someone who did NOT personally witness it. Note there may be SEVERAL separate accounts naming
different culprits; judge each person by their own chain back to whoever actually saw something.
\end{quote}

\paragraph{Flawed definition (the named-teller trap).}
The original extraction prompt defined \emph{observed} = saw it; \emph{told} = a specific named person
told it; \emph{rumor} = heard it as a crowd/market rumor. Because ``told'' keys on \emph{having a named
teller} rather than on \emph{whether the teller established the fact}, every relayed claim with a known
teller classifies as testimony, collapsing the testimony/hearsay distinction at depth $\geq 2$
(verbatim prompt in \texttt{audit\_gold.py}).

\paragraph{Gold audit.}
A per-agent audit of the controlled chains (Claude-Opus as auditor, majority-of-2 with tie-break)
confirmed the gold source types are correct; under the flawed definition the models' ``errors'' exactly
track the definition (they faithfully report the named teller), which is what identified the wall as a
prompt artifact rather than a capability limit.

\section{Reproducibility}
\label{app:repro}

The full release (code, frozen scenarios, verbatim prompts, row-level outputs) is browsable at \url{https://github.com/Jiajun-Jiao/launderbench}.

Environment: \texttt{OPENAI\_API\_KEY} / \texttt{ANTHROPIC\_API\_KEY} / \texttt{GOOGLE\_API\_KEY} exported; local open models via
Ollama. All commands run from the repository root; outputs land in \texttt{outputs/society/}. The
primary scenario set for the main citation-panel run is
\texttt{citation\_scenarios\_100\_v2.jsonl} (100 weak-origin scenarios + 96 model-gated matched
\texttt{solid} controls); the Table~\ref{tab:handoff} handoff runs, the format controls, and the
mixed-graph probe use it as well. Mechanism-ladder and worked-example diagnostics use the
canonical-30 \texttt{citation\_scenarios\_v2.jsonl} panel.

\begin{table}[t]
\centering
\scriptsize
\setlength{\tabcolsep}{2.5pt}
\begin{tabular}{ll}
\toprule
\textbf{result} & \textbf{script (experiments/)} \\
\midrule
typing gradient & \texttt{provenance\_complexity.py} \\
clean-chain control & \texttt{provenance\_trace.py} \\
citation panel & \texttt{research\_citation\_multi.py} \\
mechanism ladder & \texttt{provenance\_mechanism\_diagnostic.py} \\
scenario synthesis & \makecell[l]{\texttt{citation\_scenarios.py},\\\texttt{citation\_solid.py}} \\
seeded extension & \makecell[l]{\texttt{benchmark\_seed\_adapter.py},\\\texttt{citation\_seeded.py}} \\
native controls & \makecell[l]{\texttt{external\_native\_prepare.py},\\\texttt{external\_native\_eval.py}} \\
caveat pilot & \texttt{caveat\_laundering.py} \\
stress tests & \makecell[l]{\texttt{provenance\_pipeline.py},\\\texttt{provenance\_relay.py},\\\texttt{provenance\_pressure.py}} \\
type-once-carry & \makecell[l]{\texttt{provenance\_handoff.py},\\\texttt{handoff\_condition\_analyze.py}} \\
definition audit & \texttt{audit\_gold.py} \\
judge robustness & \texttt{judge\_robustness\_audit.py} \\
reasoning/CoT rows & \texttt{research\_citation\_multi.py} (\texttt{COT=1}) \\
LangChain replication & \texttt{framework\_langchain.py} \\
gpt-researcher runs & \makecell[l]{\texttt{framework\_gptr.py},\\\texttt{gptr\_mock\_search.py}} \\
real-web case bank & \texttt{real\_web\_pilot.py} \\
human audits & \makecell[l]{\texttt{citation\_human\_audit\_score.py},\\\texttt{scenario\_audit\_score.py}} \\
\bottomrule
\end{tabular}
\caption{Result-to-script map. The repository's \texttt{REPRODUCE.md} lists the exact command for
every number in this paper.}
\label{tab:repro}
\end{table}

The exact one-line invocations for the Table~\ref{tab:repro} scripts are listed in \texttt{REPRODUCE.md}; the main panel run sets
\texttt{SCENARIOS} to \texttt{citation\_scenarios\_100\_v2.jsonl}.

\paragraph{Artifact completeness.}
The original ExpertQA Opus baseline retains an aggregate summary but not its row-level file; the
21/28 valid-answer reinterpretation is documented by the contemporaneous refusal audit rather than
recomputed from baseline rows. For handoff, a schema-v2 replication supersedes the original
typed-handoff result: all 400 weak-origin and 384 matched-solid carrier answers, briefings, fields,
and verdicts are preserved for offline re-adjudication. The runner also records attempted and valid
denominators.

\paragraph{Model identifiers, collection dates, decoding.}
All commercial-API data were collected 2026-06-28 through 2026-07-29 with the providers' model IDs
verbatim as follows (the wrappers did not return a separate resolved snapshot ID, so collection
dates bound provider-side alias resolution): OpenAI \texttt{gpt-4o-mini}, \texttt{gpt-4o}, \texttt{gpt-4.1-mini},
\texttt{gpt-4.1}, \texttt{o3}, \texttt{o4-mini}, \texttt{gpt-5-mini}, \texttt{gpt-5.4-mini},
\texttt{gpt-5.4-nano}, \texttt{gpt-5.5}; Anthropic \texttt{claude-haiku-4-5},
\texttt{claude-sonnet-4-6}, \texttt{claude-opus-4-8}; Google \texttt{gemini-2.5-flash} (gate
replication: \texttt{gemini-2.5-pro}; frontier probe and third scenario author:
\texttt{gemini-3.1-pro-preview}). Open-weight models ran locally via Ollama under the exact
tags \texttt{qwen2.5:7b-instruct}, \texttt{qwen2.5:32b-instruct}, \texttt{llama3.1:8b},
\texttt{mistral:7b}, \texttt{gemma2:9b}, and \texttt{qwen3:30b-a3b-instruct-2507-q4\_K\_M}.
Open-weight models are used per their technical
reports~\citep{qwen2025qwen25,qwen2025qwen3,grattafiori2024llama3,jiang2023mistral,gemmateam2024gemma2};
commercial models follow the providers' official model
documentation~\citep{openai2026docs,anthropic2026docs,google2026docs}. Local inference ran through
Ollama 0.11.8~\citep{ollama2023} on one NVIDIA RTX 5090 (32\,GB) under Python 3.13, totaling on the
order of tens of GPU-hours across pilot and final open-weight runs. The
canonical 100-scenario panel was collected 2026-07-02/03; every row-level record ships with the
release. The answer harness shuffles each scenario's five pages with a fixed-seed RNG
(\texttt{SEED}=20260630) before rendering, so page order is randomized per scenario but exactly
reproducible; the root's position is therefore not fixed. In the main answer harness, OpenAI non-reasoning models and Gemini 2.5 Flash run with temperature 0; OpenAI reasoning
models and Anthropic use provider-default temperature. No reasoning-effort override is set: OpenAI
reasoning models receive \texttt{max\_completion\_tokens=25000}, Anthropic
\texttt{max\_tokens=1024}, while non-reasoning OpenAI and Gemini leave the output cap at the provider
default. Gemini uses no \texttt{thinkingConfig} override (the API-default dynamic thinking budget).
The decoding-variance supplement uses temperature 0.7 on the two supported OpenAI
models. Refusal or format failures are recorded as \texttt{api\_error} rows and
excluded from denominators (one such row in the whole canonical panel, on
\texttt{qwen2.5:7b-instruct}).

Framework runs are pinned: LangChain 1.3.11 / langchain-core 1.4.8 / langchain-openai 1.3.3 /
langchain-anthropic 1.4.8 (venv \texttt{.venv-frameworks/}); Haystack 2.30.2 and LlamaIndex 0.14.23
(venv \texttt{.venv-agentfw/}); gpt-researcher 0.15.1 (venv
\texttt{.venv-gptr/}), \texttt{RETRIEVER=custom} against the local mock endpoint, FAST/SMART/STRATEGIC
LLM all set to the target model, embeddings at the framework default.

\paragraph{Prompt development and freezing.}
The canonical answer instruction, gate prompts, and judge rubric were developed iteratively on the
earlier 30-scenario pilot set and small smoke subsets, and were then held fixed for every row of the
canonical 100-scenario panel and all replications; scenario sets are versioned, and the released
files are the frozen v2 instruments. No prompt was tuned per model: every evaluated model receives
the identical canonical instruction, and all instruction variants (neutral, deleted-instruction,
framework-default) appear only as labeled ablations. The typing, handoff, mechanism, and judge
prompts are released verbatim (Appendix~\ref{app:prompts}), and no hyperparameter search was
performed beyond the fixed decoding settings above.

\paragraph{Cost.}
Total commercial-API spend across all experiments in this paper is on the order of a few hundred US
dollars; a single 100-scenario, three-condition panel costs roughly US\$1--10 per model, and each full
strong-judge relabeling of the canonical panel (2{,}072 answers) cost US\$7--9. Open-weight rows ran
locally via Ollama at no API cost.

\section{Full Result Tables}
\label{app:tables}

\paragraph{Alternative explanations for the cohort split.}
The argument would be weak if benchmark construction or the judge created the cohort split.
Neither appears to. Two non-Claude authors change absolute difficulty but not cohort membership;
a Gemini validator confirms every construct and solid criterion, with six prior-freeness dissents
leaving survivor rates within .03; and four blinded human annotators majority-pass 39/40 audited
scenarios (40/40 for the hypothesis-unaware three) while recovering all 40 gold origins.
Removing every item with any dissent moves no main rate by more than .022. Two stronger judges agree with the production
judge on 0.978--0.983 of 2,072 binary decisions, and the blinded human audit reaches
$\kappa{=}1.00$ on the established versus non-established distinction.

Nor does the split reduce to generic reasoning effort. A free-form reasoning field leaves
GPT-4o-mini and GPT-4.1-mini at 0.89 and 0.63 laundering, whereas provenance-directed instructions
and explicit types sharply reduce it. Reasoning models are also mixed: within the same low-cost tier, gpt-5-mini reaches zero while
o4-mini remains at 0.32. We therefore read the result as provenance-state use, not a general
reasoning or price law.

Framework scaffolding does not repair the missing decision either. With the same pages,
GPT-4o-mini launders 0.97--0.99 under generic Haystack, LangChain, and LlamaIndex
prompts~\citep{pietsch2019haystack,chase2022langchain,liu2022llamaindex} while still
accepting every solid control. In a mock-search agent loop, planners never retrieve the weak origin; Haiku nonetheless refrains
once the supplied context shows dependence.
Planning failure can amplify laundering but is neither necessary nor sufficient for the model
difference. Synthetic framing is the last alternative: treating placeholder domains as genuine increases
acceptance in one cautious Opus solid cell while leaving its weak-origin rate at zero: the
framing makes the estimate conservative rather than manufacturing the split.

\paragraph{Source typing.}
Table~\ref{tab:typing} gives the naturalistic-prose typing gradient summarized in
Section~\ref{sec:exp_typing}. Table~\ref{tab:cleanchain} gives the clean-chain control
(\texttt{experiments/provenance\_trace.py}; one chain, depths 1--6, four chains per depth, 60
gold-hearsay classifications per condition), which runs on all twelve panel snapshots and holds
everything fixed except whether the propagation graph must be recovered from text. Four snapshots
score 1.00 in both renderings and Gemini Flash misses a single hop-depth-two item under the edge
list; Qwen2.5-32B and Qwen3-30B score 0.78 in both by calling depth-one testimony hearsay, never the
reverse. Among the five that break, the sign of the
prose-to-edges difference splits them: positive for GPT-4.1 and Haiku, near zero for GPT-4o-mini,
and negative for GPT-4.1-mini and GPT-4o, whose accuracy under the explicit edge list falls with
depth (GPT-4o 1.00 at depths zero and one, then .65/.56/.42/.38/.25) while its prose rendering
stays at 1.00 throughout. The instrument's original expectation was the positive sign only, and
two rows invert it. The two instruments differ in generator,
rendering, competing-chain count, and depth, so they bound the task from two sides rather than
forming a matched contrast; only the prose-versus-edges split inside the clean-chain control is an
isolation.

\begin{table}[t]
\centering
\scriptsize
\setlength{\tabcolsep}{3pt}
\begin{tabular}{lccc}
\toprule
\textbf{model} & \makecell{\textbf{prose}\\(reassemble)} & \makecell{\textbf{edges}\\(supplied)} & \makecell{\textbf{over-attr.}\\\textbf{prose; edges}} \\
\midrule
GPT-4.1            & 0.85 & 0.99 & 15; \phantom{0}1 \\
Claude Haiku 4.5   & 0.93 & 1.00 & \phantom{0}8; \phantom{0}0 \\
GPT-4o-mini        & 0.59 & 0.62 & 44; 41 \\
\addlinespace
Claude Opus 4.8    & 1.00 & 1.00 & \phantom{0}0; \phantom{0}0 \\
Claude Sonnet 4.6  & 1.00 & 1.00 & \phantom{0}0; \phantom{0}0 \\
GPT-5.5            & 1.00 & 1.00 & \phantom{0}0; \phantom{0}0 \\
o3                 & 1.00 & 1.00 & \phantom{0}0; \phantom{0}0 \\
Gemini 2.5 Flash   & 1.00 & 0.99 & \phantom{0}0; \phantom{0}1 \\
\addlinespace
Qwen2.5-32B        & 0.78 & 0.78 & \phantom{0}0; \phantom{0}0 \\
Qwen3-30B          & 0.78 & 0.78 & \phantom{0}0; \phantom{0}0 \\
\addlinespace
GPT-4.1-mini       & 0.81 & \textbf{0.55} & 20; \textbf{49} \\
GPT-4o             & 1.00 & \textbf{0.73} & \phantom{0}0; \textbf{29} \\
\bottomrule
\end{tabular}
\caption{Clean-chain control: source-type accuracy when the model must reassemble the chain from
shuffled prose versus when the chain is supplied as an explicit edge list. Over-attribution counts
gold-hearsay agents called \emph{testimony}, out of 60 per cell. Rows are grouped by the sign of
the prose-to-edges difference: supplying the structure repairs the first group, does not repair
GPT-4o-mini, is irrelevant to the middle groups, and \emph{harms} the last two.}
\label{tab:cleanchain}
\end{table}

\begin{table}[t]
\centering
\scriptsize
\setlength{\tabcolsep}{3pt}
\begin{tabular}{lccc}
\toprule
\textbf{model} & \makecell{\textbf{testim.}\\\textbf{(d1)}} & \makecell{\textbf{hearsay}\\\textbf{(d$\geq$2)}} & \makecell{\textbf{hearsay$\rightarrow$}\\\textbf{testim.}} \\
\midrule
o3, Sonnet-4.6, Opus-4.8 & 1.00 & 1.00 & 0.00 \\
GPT-4.1                  & 0.80 & 0.68 & 0.33 \\
Claude-Haiku-4.5         & 0.80 & 0.62 & 0.38 \\
GPT-4.1-mini             & 0.90 & 0.45 & 0.55 \\
\midrule
\makecell[l]{Qwen2.5-7B/32B, Llama-3.1-8B,\\Mistral-7B, Gemma2-9B} & 0.00--0.20 & 0.80--1.00 & 0.00--0.20 \\
\bottomrule
\end{tabular}
\caption{Source-typing accuracy under naturalistic prose with competing chains ($K{=}2$). Each
model receives five seeded repetitions: 60 classifications total
(10 direct-observation, 10 testimony, 40 hearsay); missing or malformed predictions count as
incorrect. Models
added later for the citation panel (the GPT-4o family, GPT-5.5, Gemini) are not in this wing; the
mechanism probes cover low-cost-tier typing directly.}
\label{tab:typing}
\end{table}

\paragraph{Low-cost reasoning tier on the canonical panel.}
Table~\ref{tab:reasoning} gives the per-model rates behind the reasoning-effort argument
above.

\begin{table}[t]
\centering
\scriptsize
\setlength{\tabcolsep}{2.5pt}
\begin{tabular}{lccccc}
\toprule
\textbf{model} & \texttt{trace}$\downarrow$ & \textbf{95\% CI} & \texttt{flag.}$\downarrow$ & \texttt{solid}$\uparrow$ & \textbf{discr.} \\
\midrule
o4-mini       & \textbf{0.32} & [.24,\,.42] & 0.00 & 1.00 & $+$0.67 \\
gpt-5-mini    & \textbf{0.00} & [0,\,.04]   & 0.00 & 1.00 & $\mathbf{+1.00}$ \\
gpt-5.4-mini  & \textbf{0.07} & [.03,\,.14] & 0.00 & 0.86 & $+$0.79 \\
gpt-5.4-nano  & 0.07          & [.03,\,.14] & 0.00 & 0.70 & $+$0.62 \\
\bottomrule
\end{tabular}
\caption{Low-cost reasoning and newest-mini rows on the 100-item canonical panel. Discrimination is
paired over the 96 matched controls, as in Table~\ref{tab:panel}.}
\label{tab:reasoning}
\end{table}

\paragraph{Topology breakdown.}
\texttt{trace} laundering by topology, as linear~(/34) / star~(/33) / fake corroboration~(/33):
GPT-4o-mini 34/29/33; GPT-4.1-mini 32/14/28; GPT-4.1 33/7/22; Qwen3-30B 22/12/19; o3 4/0/1.
\texttt{linear} is the highest-laundering condition for every row and \texttt{star} the lowest or
tied for every panel model, with \texttt{fake corroboration} between: two seemingly independent chains conceal the shared root better than a visible hub. We treat
topology differences as robustness evidence, not a causal effect: the cells contain different authored
scenarios.

\paragraph{Gate-family re-validation.}
\texttt{experiments/citation\_regate.py} re-runs the prior-freeness and construct-validity gates
(validators V2 and V3 in the script) plus the solid gate over the canonical 100 with Gemini 2.5 Pro (a validator outside the evaluated
panel) under the original prompts and
schemas, then re-tabulates every model's rates on the survivor subset from the released row-level
records (no new panel calls). Verdicts: 94/100 \texttt{trace} scenarios and 90/96 \texttt{solid}
pairs survive; all six exclusions are prior-freeness disagreements in which Gemini reports a
\emph{false} prior on the invented claim; no entity is recognized as real, and no construct-validity
or solid criterion fails on any scenario. Survivor-subset rates for all 22 recorded model rows move by at
most 0.03 with every ordering preserved
(\texttt{outputs/society/}\allowbreak\texttt{citation\_regate\_gemini25pro.json}).

\paragraph{Closed-book prior probe.}
The prior-freeness gate rests on one validator, so \texttt{experiments/closed\_book\_prior.py}
probes every panel model directly: each of the 100 canonical questions is asked with \emph{no}
retrieved pages under an instruction that forces a single best guess, scored by a deliberately
generous lexical hit rule (every number in the target value, format- and number-word-insensitive;
otherwise any distinctive value token absent from the question itself). Every stored hit was
inspected by hand. Across the 19 non-author models (1{,}900 probes) there are 11 hits, 0--2 per
model: nine are single generic-token coincidences (``fermented'' alone accounts for five models on
one scenario, plus ``dried,'' ``forfeits'' twice, and a lone ``three''), and gpt-5.4-nano
contributes the only two plausible-value re-guesses (a village population, 214, stated confidently;
a prodigy age, nineteen, offered as an explicit guess), both on items its own panel verdicts
\emph{hedge}: no panel number rests on parametric availability. The scenario author (Claude
Opus) is the informative exception at 4/100: three hits re-derive its own invented values (1483;
47 metres; 4 minutes 12 seconds), each labeled unverified or fictional in the same answer; author
self-consistency is detectable but epistemically inert, and the author's panel row is 0/100
laundering. Decline-style language despite the forced-guess instruction ranges from 0/100
(GPT-4o-mini, o3, o4-mini, gpt-5-mini) to 79/100 (Sonnet).

\paragraph{Instruction-neutrality ablation.}
The canonical task prompt asks the model to ``state how well established the answer is'' and to
``be accurate about the strength of the evidence.'' Two ablations weaken this on the canonical
panel (\texttt{INSTR=} in \texttt{research\_citation\_multi.py}; judge instrument unchanged; run
for the cohort extremes, the two heaviest launderers and two zero-rate models):
\emph{no-accuracy} drops only the accuracy sentence, \emph{neutral} drops both, leaving a bare
answer-the-question prompt. \texttt{trace} laundering: GPT-4o-mini 96 (canonical) / 97
(no-accuracy) / 95 (neutral) per 100; GPT-4.1-mini 74 / 77 / 82. Under \emph{neutral}, GPT-5.5 and
Claude Sonnet stay at 0/100 (Sonnet returning an explicit traced-weak-origin verdict on 100/100
items), while \texttt{solid} acceptance holds or \emph{rises} (GPT-5.5 95/96, discrimination
$+0.99$; Sonnet 76$\rightarrow$94/96, $+0.98$), so part of Sonnet's conservatism is itself
instruction-induced rather than intrinsic. The instruction is
therefore charitable rather than load-bearing: deleting it worsens the launderers (or leaves them
at ceiling) and does not produce the frontier zeros. It \emph{is} load-bearing for GPT-4o-mini's
use of the explicit flag (\texttt{flagged} misses 4$\rightarrow$28/100 under \emph{neutral};
GPT-4.1-mini is instruction-insensitive at 1$\rightarrow$1/100), reproducing the stack-default flag
misses (20/14/5 per 100): with the flag verifiably
in the input, missing it under a bare prompt is prompt sensitivity, not missing evidence.

\paragraph{Handoff format controls.}
\texttt{experiments/provenance\_handoff\_prose.py} adds three arms on the canonical 100, reusing the
Table~\ref{tab:handoff} typer fields verbatim: \emph{prose-warning} (Opus writes a prose-only
briefing under an explicit instruction to state how the value is grounded; a lexical check finds the
warning in 100/100 briefings), \emph{field-vs-prose} (the carrier answers from its own plain,
typically confident briefing plus Opus's typed field, which conflict), and \emph{templated}, a
content-matched control for the typed arm: the same stored Opus briefing and the same stored typed
field, with the field rendered by fixed code into one English sentence
(``The key value is \ldots{} Provenance: \ldots{} Independently verified: \ldots{} Basis: \ldots'')
instead of JSON, so no model writes the record and no caveat can enter it.
All four carriers launder
0/100 in each arm (0/400 pooled each), with nearly every answer returning an explicit
traced-weak-origin verdict. A schema-v2 rerun stores all 400 typed weak-origin answers; the matched
\texttt{solid} arm stores 384 more and yields 96/96 establishments for every typed carrier. The
prose-warning channel remains weak-safe but establishes 85/96, 86/96, 86/96, and 87/96 solids
(paired discrimination $.885$--$.906$); its generated briefings sometimes add caveats beyond the
independence judgment. The templated arm instead reproduces the typed arm exactly: 96/96 solids for
every carrier and paired discrimination $1.000$ $[1.000, 1.000]$. Holding the decision's content
fixed and varying only its encoding therefore changes nothing, which locates the prose deficit in
regeneration rather than in the absence of a schema. Per-carrier \emph{self-carry} splits (Table~\ref{tab:handoff} rows, in
carrier order): laundering 0/23, 0/55, 0/97, 0/92 when the carrier's own field says weak/circular,
and 73/77, 44/45, 3/3, 8/8 otherwise. Together with prose-only self-handoff at 0.76--0.96, these results show that the operative
variable is whether the handoff carries the provenance conclusion unaltered: default prose drops it,
instructed prose regenerates it and keeps only weak-origin restraint, and a verbatim record, as JSON
or as a templated sentence, carries a stable, machine-checkable license in both directions while
surviving conflicting prose.

\paragraph{Low-cost-typer replication and measured cost.}
Replacing the Opus typer with \emph{gpt-5-mini} (same seed, typed arm only) leaves the protocol
intact: its field flags weak/circular provenance on 100/100 scenarios and all four carriers stay at
0/100 (0/400 pooled). Cost is measured from API-reported token usage
(\texttt{experiments/}\allowbreak\texttt{typer\_cost\_\{calibration,report\}.py}; list prices as
of 2026-07-07).
Opus-direct answering costs \$12.8 per 1{,}000 queries (850 in / 341 out tokens per query); typing
once costs \$11.3 per 1{,}000 scenarios (1{,}156 in / 220 out) and a carrier consumption \$0.09 per
1{,}000 (283 in / 77 out; the typed bundle is $3\times$ smaller than the raw pages, which also
shrinks downstream context). Because a typed briefing is shorter than a direct Opus answer, typed
handoff undercuts Opus-direct already at \emph{one} consumption per bundle (\$11.4 vs.\ \$12.8 per
1{,}000; break-even $k^\ast{=}0.89$) and amortizes to $10.5\times$ cheaper at $k{=}10$
consumptions. The gpt-5-mini typer cuts typing to \$1.4 per 1{,}000 (714 in / 604 out including
reasoning tokens), putting the whole protocol at \$1.5 per 1{,}000 single-consumption queries,
$8.7\times$ below Opus-direct. These ratios use Opus-direct as the comparator; they do not show that
a handoff is cheaper or simpler than direct answering by gpt-5-mini, which is already an eligible
low-laundering answerer. The operational case is strongest when one provenance decision is audited
or reused by multiple downstream consumers.

\paragraph{Typer-eligibility sweep.}
Carriers execute the carried field near-deterministically: pooled over every arm in which the
field is correct (the self-carry correct-field cells and all three full typer runs), laundering is
0/1{,}303, versus 287/297 (0.97) where it is wrong. A deliberately weak typer confirms the mapping
causally: gpt-4.1-mini types 59/100 (every mistype a confident
\texttt{independent\_primary\_sources}), and the four carriers then launder 39--41/100 each,
almost exactly the mistyped set (159/164 on field-wrong scenarios vs.\ 0/236 on field-correct;
predicted $41\times0.97\approx40$). Per-scenario typing correctness is therefore a sufficient
statistic for the protocol outcome, and typer eligibility can be scored with the typing instrument
alone: \texttt{experiments/typing\_sweep.py} runs the verbatim briefing+field prompt on
\texttt{trace} (gold: weak/circular) and \texttt{solid} (gold: independent-primary), machine-checked
against gold with no carriers and no judge (Table~\ref{tab:typersweep}).

\begin{table}[t]
\centering
\scriptsize
\setlength{\tabcolsep}{4pt}
\begin{tabular}{lccc}
\toprule
\textbf{candidate typer} & \texttt{trace}$\uparrow$ & \texttt{solid}$\uparrow$ & \textbf{t-disc} \\
\midrule
GPT-4o-mini        & 29/100  & 96/96 & 0.29 \\
GPT-4.1-mini       & 61/100  & 96/96 & 0.61 \\
GPT-4o             & 97/100  & 96/96 & 0.97 \\
GPT-4.1            & 90/100  & 96/96 & 0.90 \\
Gemini 2.5 Flash   & 100/100 & 96/96 & \textbf{1.00} \\
\addlinespace
o4-mini            & 100/100 & 96/96 & \textbf{1.00} \\
gpt-5.4-mini       & 100/100 & 96/96 & \textbf{1.00} \\
gpt-5.4-nano       & 98/100  & 54/96 & 0.54 \\
gpt-5-mini         & 100/100 & 96/96 & \textbf{1.00} \\
\addlinespace
o3                 & 100/100 & 96/96 & \textbf{1.00} \\
GPT-5.5            & 100/100 & 96/96 & \textbf{1.00} \\
Claude Haiku 4.5   & 100/100 & 96/96 & \textbf{1.00} \\
Claude Sonnet 4.6  & 100/100 & 96/96 & \textbf{1.00} \\
Claude Opus 4.8    & 100/100 & 96/96 & \textbf{1.00} \\
\addlinespace
Qwen2.5-7B         & 95/100  & 32/96 & 0.28 \\
Qwen2.5-32B        & 91/100  & 10/96 & 0.01 \\
Qwen3-30B          & 93/100  & 14/96 & 0.08 \\
Llama-3.1-8B       & 67/100  & 56/96 & 0.25 \\
Mistral-7B         & 100/100 & 1/96  & 0.01 \\
Gemma2-9B          & 95/100  & 34/96 & 0.30 \\
\bottomrule
\end{tabular}
\caption{Typer-eligibility sweep: the verbatim briefing+field typer instrument, machine-scored
against gold (\texttt{trace}: field says weak/circular; \texttt{solid}: independent-primary);
t-disc $=$ trace rate $+$ solid rate $-$ 1. Predicted downstream laundering per 100 consumptions
$\approx$ mistyped \texttt{trace} $\times$ 0.97, validated on three measured anchor runs: Opus
100/100 fields $\rightarrow$ 0/400, gpt-5-mini 100/100 $\rightarrow$ 0/400, GPT-4.1-mini 59/100
$\rightarrow$ 159/400 (59/100 is that run's own field accuracy; this sweep instrument scores the
same model 61/100).}
\label{tab:typersweep}
\end{table}

\paragraph{Who can type?}
Nine models are fully eligible (t-disc 1.00): the three Claude models, o3, GPT-5.5, gpt-5-mini,
gpt-5.4-mini, o4-mini, and Gemini 2.5 Flash, so every tested provider family has an eligible low-cost typer.
Three readings. First, the diagnostic--generation gap is panel-wide: o4-mini (types 100/100,
launders 32/100 raw), Gemini 2.5 Flash (100/100 vs.\ 29/100), and GPT-4o (97/100 vs.\ 72/100) perform
better when the prompt explicitly requests the intermediate provenance label. Second, the \texttt{solid} column keeps the
table honest: Mistral-7B types \texttt{trace} at 100/100 by calling \emph{everything} weak
(\texttt{solid} 1/96), and the Qwen family behaves similarly (10--32/96); a blanket-weak typer
would suppress laundering while poisoning every well-sourced answer, the typer-side analogue of
blanket caution. Third, the only models that fail the \texttt{trace} side outright are the low-cost
OpenAI non-reasoning minis (GPT-4o-mini 29/100, GPT-4.1-mini 61/100), the cohort that also
launders most. This correlation is descriptive across the tested configurations; the experiment
does not isolate a training-recipe variable.

\paragraph{Statistical treatment.}
All models answer the same 100 \texttt{trace} scenarios, so tier comparisons use exact McNemar
tests on paired per-scenario outcomes (\texttt{experiments/panel\_stats.py}): GPT-4o-mini vs.\
GPT-4o 24/0 discordant scenarios ($p{=}1.2{\times}10^{-7}$), GPT-4.1-mini vs.\ GPT-4.1 15/3
($p{=}7.6{\times}10^{-3}$), and every ${\geq}0.53$ launderer vs.\ every zero-rate model at
$p{\leq}2.3{\times}10^{-16}$ (GPT-4o-mini vs.\ GPT-5.5: 96/0, $p{=}2.5{\times}10^{-29}$).
Discrimination scores use the 96 paired \texttt{trace}/\texttt{solid} items and carry
percentile-bootstrap 95\% intervals (10{,}000 paired resamples): GPT-4o-mini $+0.03$
$[0.00, {+}0.07]$, GPT-4.1 $+0.38$ $[{+}0.28, {+}0.48]$, Gemini 2.5 Flash $+0.71$
$[{+}0.61, {+}0.79]$, and Claude Haiku $+0.98$ $[{+}0.95, {+}1.00]$. GPT-4o-mini's interval includes
zero; its high laundering rate is not evidence of positive discrimination. Because scenarios share a generation pipeline
balanced over 20 domains (per-domain rates in Table~\ref{tab:domains}), every \texttt{trace} rate is also recomputed under a domain-clustered
bootstrap (resampling the 20 domains with replacement): clustered bounds sit within 0.02--0.03 of
the scenario-level Wilson bounds for the OpenAI rows (GPT-4o-mini $[.92,.99]$ vs.\ Wilson
$[.90,.98]$) and widen most for Gemini 2.5 Flash ($[.16,.43]$ vs.\ $[.21,.39]$), without changing
any ordering or significance statement.

\paragraph{Decoding-variance supplement.}
For the two heaviest launderers that accept an explicit sampling temperature, three
temperature-0.7 repetitions yield mean \texttt{trace} laundering of 0.90 (GPT-4o-mini) and 0.78
(GPT-4.1-mini), with run-to-run standard deviation below 0.01. At the opposite pole, the five
near-zero snapshots were rerun three times each under provider-default decoding on a stratified
30-scenario subset (seed 20260729; ten per topology from the 96 matched scenarios; \texttt{trace}
and \texttt{solid}; 900 answers, none invalid). Every repetition reproduces the zeros:
\texttt{trace} stays 0/30 across all twelve GPT-5.5, Haiku, Sonnet, and Opus repetitions, o3
posts 2, 1, and 0 against 2/30 in the canonical pass on the same items, and \texttt{solid}
acceptance stays within three items of the canonical draw for every snapshot (o3 and Haiku 30/30
throughout, GPT-5.5 28--29, Opus 26--28, Sonnet 22--25). The near-zero cells are therefore not
single-draw artifacts. Neither supplement is folded into the scenario-level confidence
intervals.

\begin{table}[t]
\centering
\scriptsize
\setlength{\tabcolsep}{2.5pt}
\begin{tabular}{lclc}
\toprule
\textbf{domain (abridged)} & \textbf{rate} & \textbf{domain (abridged)} & \textbf{rate} \\
\midrule
regional dish origin      & .97 & monument height          & .70 \\
named theorem credit      & .93 & word/phrase etymology    & .70 \\
record time / margin      & .90 & geographical superlative & .67 \\
battle/treaty year        & .88 & demographic statistic    & .63 \\
flag/crest symbol         & .87 & astronomical body        & .63 \\
town founding year        & .83 & first-person feat        & .63 \\
invention year            & .77 & rainfall/temperature     & .60 \\
saying attribution        & .70 & material property        & .58 \\
music premiere            & .70 & historical birthplace    & .44 \\
game rule/scoring         & .70 & animal biology           & .43 \\
\bottomrule
\end{tabular}
\caption{Per-domain \texttt{trace} laundering rate pooled over the six heaviest launderers
(GPT-4o(-mini), GPT-4.1(-mini), Qwen2.5-32B, Qwen3-30B; 4--6 scenarios per domain,
\texttt{experiments/panel\_stats.py}): laundering appears in every one of the 20 domains, with
moderate concentration (0.43--0.97).}
\label{tab:domains}
\end{table}

\paragraph{Laundering rationale analysis.}
For every answer scored \texttt{established\_fact} on a \texttt{trace} scenario, we categorize the
model's stated rationale by lexical markers (first match): \emph{multiplicity} (explicit appeal to
agreement across sources: ``multiple/several sources'', ``consistently across'', ``corroborated'',
``independent''), \emph{establishment} (vague ``well/widely established or documented'' with no source
count), \emph{authority} (defers to a downstream record), \emph{bare} (no rationale). Across the six
non-reasoning launderers, \textbf{79\%} of pooled laundered answers explicitly invoke source
\emph{multiplicity} (counting echoes rather than tracing them; 84\% across the four OpenAI
launderers), and the heaviest launderer (GPT-4o-mini) does so on 91\% (Table~\ref{tab:error}). Script:
\texttt{experiments/laundering\_error\_analysis.py}.

\paragraph{Origin diversity.}
The canonical origins span five weak-source forms: an anonymous forum guess, a personal-blog hunch,
an unsourced anonymous wiki edit, a content-farm/SEO article, and labelled fiction or satire later
taken as fact ($n{=}20/19/20/20/21$). They provide lexical and genre diversity within the shared
single-origin construct.

\paragraph{Is there a terminal-node shortcut?}
In this release, \texttt{trace} bundles contain exactly one page that cites nothing (the origin) and
\texttt{solid} bundles three, so a degenerate ``count the terminal pages'' heuristic would separate
the conditions without following any citation edge. Two observations characterize this concern,
and the mixed-graph probe below tests a related counting behavior without identifying the strategy
used in the raw-answer task. First, the cue coincides with the construct here
(the terminals of a converging citation graph \emph{are} its independent lineages); only mixed
graphs (partly independent, partly circular) can pull them apart. Second, the launderers do not
exploit the cue even when it is mechanically
trivial: with the full adjacency handed over explicitly, GPT-4o-mini still launders 22/30
(Figure~\ref{fig:mechanism}, \emph{graph} arm), and its stated rationales appeal to source
multiplicity, not terminal counts (Table~\ref{tab:error}).

\paragraph{Mixed-graph counting probe.}
For 30 topology-balanced scenarios we compose, purely from already-gated material, a \emph{mixed}
bundle: the \texttt{trace} bundle minus one in-degree-0 echo, plus one primary record lifted from
the scenario's matched \texttt{solid} control (renamed P6). Thirteen models (the nine eligible
typers and the four carriers) are asked to count the independent primary sources for the value and
to \emph{name} them, machine-scored against gold with no judge
(\texttt{experiments/mixed\_graph\_probe.py}). Gold counts are 0 (\texttt{trace}) / 1 (mixed) / 3
(\texttt{solid}); a terminal counter would answer 1/2/3 and name the terminal pages.
Table~\ref{tab:mixedgraph}: no model shows the counting pattern (the exact terminal-set signature
appears on 59 of 780 dissociating probes, concentrated in the two zero-refusers below), and every
model identifies the injected primary in the mixed bundle nearly perfectly (25--30/30): partial
corroboration is read correctly when it exists. The discriminating condition is \texttt{trace},
where the correct answer is \emph{zero}. The zero-rate models return it (o3, GPT-5.5, Sonnet, and
Opus at 30/30; Haiku 27/30), whereas GPT-4o-mini manages 3/30: on 27 items it promotes something
to primary status, the anonymous origin itself on 14 (a bibliographic reading of ``primary'' that
the instruction's establishes-the-value definition excludes: another instance of definition
sensitivity) and a confident echo on 13;
gpt-5.4-mini shows the same zero-refusal at 16/30 despite its 7/100 panel row, making the probe an
early-warning diagnostic. \texttt{solid} deviations are mild and
conservative (answering 2 by merging related records; 21--30/30). In this explicit counting frame,
the laundering disposition reappears as \emph{refusal to return an empty evidence set}, not as
terminal arithmetic. This does not establish the strategy used for ordinary answers; the matched
rewiring probe below provides a direct check. One deployment note: the typed-field enum has no label for a
graph with exactly one primary (\texttt{independent\_primary\_sources} requires ${\geq}2$), so
carrying mixed-graph verdicts needs a one-line schema extension (a count field or a
\texttt{single\_primary\_source} value), which we leave to the release.

\begin{table}[t]
\centering
\scriptsize
\setlength{\tabcolsep}{4.5pt}
\begin{tabular}{lccc}
\toprule
\textbf{model} & \makecell{\texttt{trace}\\gold 0} & \makecell{mixed\\gold 1} & \makecell{\texttt{solid}\\gold 3} \\
\midrule
GPT-4o-mini        & 3/30  & 25/30 & 25/30 \\
GPT-4.1-mini       & 19/30 & 29/30 & 28/30 \\
GPT-4o             & 29/30 & 29/30 & 22/30 \\
GPT-4.1            & 27/30 & 30/30 & 22/30 \\
Gemini 2.5 Flash   & 29/30 & 29/30 & 27/30 \\
\addlinespace
o4-mini            & 29/30 & 29/30 & 28/30 \\
gpt-5.4-mini       & 16/30 & 30/30 & 23/30 \\
gpt-5-mini         & 28/30 & 29/30 & 28/30 \\
\addlinespace
o3                 & \textbf{30/30} & 28/30 & 25/30 \\
GPT-5.5            & \textbf{30/30} & 29/30 & 21/30 \\
Claude Haiku 4.5   & 27/30 & 29/30 & 26/30 \\
Claude Sonnet 4.6  & \textbf{30/30} & 29/30 & \textbf{30/30} \\
Claude Opus 4.8    & \textbf{30/30} & 29/30 & 29/30 \\
\bottomrule
\end{tabular}
\caption{Mixed-graph counting probe: correct independent-primary counts per condition (gold
0/1/3; a terminal counter would answer 1/2/3). The exact terminal-set signature appears on 59/780
dissociating probes overall, concentrated in the GPT-4o-mini and gpt-5.4-mini zero-refusals on
\texttt{trace}; no model follows the counting pattern across conditions.}
\label{tab:mixedgraph}
\end{table}

\paragraph{Graph-only rewiring probe.}
We construct 100 matched pairs holding the question, claim, five bodies/domains, page order, warrant
profiles, and edge, terminal, and per-page indegree counts fixed
(\texttt{experiments/graph\_only\_rewire.py}). Only the P2/P5 pointers change, switching claim paths
from one to two registry roots; first-page positions are counterbalanced (20 each). All twelve panel
snapshots
answer both arms (Table~\ref{tab:graphonly}); the blinded factorized and production judges agree on
.87--1.00 of binary verdicts per model, and all numbers below use the pre-existing production rubric.
Rewiring separates four behaviors. GPT-5.5 ($+.89$, exact McNemar $p=3.2\times10^{-27}$), o3
($+.59$, $p=3.5\times10^{-18}$), and Opus ($+.50$, $p=1.8\times10^{-15}$) shift their answer
license on the edges alone; the Opus difference is positive in every first-page stratum (.25--.85),
with domain-cluster 95\% CI [.34, .66]. GPT-4o-mini, GPT-4.1-mini, GPT-4o, and Gemini 2.5 Flash
establish nearly every
bundle in both arms and do not move. Between these poles, GPT-4.1 ($+.11$), Sonnet ($+.09$,
$p=.049$), and Haiku ($+.07$, $p=.09$) establish 53--89\% of \emph{dependent} arms. The two
open-weight rows sit just below zero, Qwen3-30B $-.07$ ($p=.065$) and Qwen2.5-32B $-.06$ ($p=.11$):
neither reaches significance under the exact test this instrument uses, so we read them as no edge
use rather than as a reversal, and note that the paired bootstrap interval excludes zero for
Qwen3-30B alone. For Haiku and
Sonnet, whose canonical \texttt{trace} rates are ${\le}.05$, this is the sharpest observation: once
every page shares the same neutral registry-entry rendering, they establish most dependent bundles, so their
canonical restraint tracks the origin's own text rather than the citation edges. Explicit
reading remains dissociated from use for the models that can read: the six snapshots recovering
exact root-ID sets on 88--100\% of arms include both poles, Opus, o3, GPT-5.5, and Sonnet as well as
Gemini and GPT-4.1, so recovery does not predict the licensing move. Recovery itself is not
universal: GPT-4o-mini scores 0, and the low-cost and open-weight rows recover the dependent arm's
single root far less often than the independent arm's pair (GPT-4.1-mini 35 vs.\ 92, GPT-4o 59 vs.\
100, Qwen2.5-32B 9 vs.\ 50, Qwen3-30B 1 vs.\ 43). The table's joint diagnostic is stricter, also requiring the declared root
count, and Haiku most often fails only that count on dependent arms (17 joint vs.\ 89 exact-ID):
one dependent-arm answer names both registry entries while calling them
``a chain of independent sources.'' Thus the controlled contrast establishes decisive edge sensitivity for
three snapshots across two providers, with smaller reliable shifts in GPT-4.1 and Sonnet; the factorial below decomposes the origin-text channel this
instrument holds fixed.

\paragraph{Google frontier probe.}
The panel's frontier tier spans two laboratories, so we additionally evaluated a Google frontier
snapshot, gemini-3.1-pro-preview (also the third scenario author of Table~\ref{tab:author}), on
both core instruments; the panel's composition was frozen before collection, so it is reported
here (collected 2026-07-29). On the canonical panel it posts 0/100 \texttt{trace}, 0/100
\texttt{flagged}, and 96/96 \texttt{solid}, paired discrimination $+1.00$. On the rewiring probe
it shifts on the edges alone: 18/100 dependent versus 99/100 independent establishments under the
production rubric ($+.81$ $[.73,.88]$, exact McNemar $p{=}8.3\times10^{-25}$, discordant pairs
81:0), with exact root recovery on 100/100 of both arms and 74 dependent-arm answers labeled
traced-weak-origin. Its factorized-versus-production agreement (.74) sits below the panel range
because it often reports the value while spelling out the single lineage, a stance the production
rubric reads as tracing rather than establishment. The edge channel is therefore not a
two-laboratory artifact: its carriers span OpenAI, Anthropic, and Google frontier snapshots.

\paragraph{Origin text $\times$ edges factorial.}
A second instrument crosses the same 100 pairs with an origin-text axis
(\texttt{graph\_only\_rewire.py build --register confessed}): only the symmetric P2/P3 root pages
change, from dated registry entries to personal notes whose bodies add ``the note's author calls
this a personal guess, with no measurement, record, or verification behind it'' (a constant 46
characters; claim sentence, remaining pages, wiring, order, and counts unchanged). The origin-text
channel moves every snapshot. With the origin self-described as a guess, no snapshot establishes
more than 4/100 in either arm: ten of the twelve establish 0/100 independent bundles (Qwen2.5-32B
1/100, Qwen3-30B 2/100) and eight of the twelve 0/100 dependent bundles (GPT-4o-mini 2/100,
Qwen3-30B 4/100, GPT-4.1-mini and Qwen2.5-32B 1/100 each). Two independent confessed guesses are
therefore almost never read as corroboration, and the two judges agree on .90--1.00 of confessed-row
binary verdicts. Dependent-column origin-text effects mirror each model's registry-row rates:
$+.97$--$+1.00$ (GPT-4o-mini, Gemini, GPT-4.1-mini, GPT-4o), $+.89$ (GPT-4.1), $+.68$--$+.72$
(Haiku, Qwen2.5-32B, Qwen3-30B), $+.53$ (Sonnet; exact McNemar
$p\le2.2\times10^{-16}$ for the Claude rows), and near zero for the three edge-sensitive snapshots whose dependent
suppression the edges already produce (Opus $.00$, GPT-5.5 $.04$, o3 $.12$). Independent-column
origin-text effects reach $+.51$ to $+1.00$ for every model that established registry-row independent
bundles. The factorial therefore decomposes this instrument's two channels: the origin's text moves
all twelve snapshots, citation edges move only GPT-5.5, o3, and Opus.

One disclosure: Opus returns
an Anthropic safety refusal on all four G001 calls (the fictional compound name collides with a
real explosive; deterministic on retry), so that pair is excluded and reported, leaving 99 pairs.
A full-panel audit finds no other refusal or error row: 4/9,600 graph-only rows (all G001), and
none among the 7,888 canonical-100 rows across all released runs.
Recorded cost: US\$5.70 over priced models. The confessed row is deliberately the salient end of
the origin-text axis: the same note appears on both root pages, next to the claim, whereas a canonical
origin hedges once, in free prose, on one page outvoted by four assertive echoes, and some
canonical origin kinds never state their weakness at all. Canonical \texttt{trace} therefore sits
between the factorial's rows, and its laundering rate measures whether a model surfaces and applies
a buried origin-text signal, consistent with the near-zero \texttt{flagged} rates. The confessed level changes the origin's self-described warrant, an explicit textual cue rather
than style alone; a style-only rewrite holding the stated warrant fixed remains untested. The
canonical contrast also still varies free-form authorship and origin phrasing, and each factorial
cell remains one generation per arm.

\begin{table}[t]
\centering
\scriptsize
\setlength{\tabcolsep}{4pt}
\begin{tabular}{lccccc}
\toprule
\textbf{model} & \textbf{$N_{\text{laund}}$} & \textbf{mult.} & \textbf{estab.} & \textbf{auth.} & \textbf{bare} \\
\midrule
GPT-4o-mini  & 96 & 91\% & \phantom{0}9\% & 0\% & 0\% \\
GPT-4.1-mini & 74 & 78\% & 22\% & 0\% & 0\% \\
GPT-4o       & 72 & 99\% & \phantom{0}1\% & 0\% & 0\% \\
GPT-4.1      & 62 & 61\% & 39\% & 0\% & 0\% \\
Qwen2.5-32B  & 70 & 77\% & 23\% & 0\% & 0\% \\
Qwen3-30B    & 53 & 58\% & 42\% & 0\% & 0\% \\
\bottomrule
\end{tabular}
\caption{Failure-mode breakdown of laundered \texttt{trace} answers: laundering overwhelmingly appeals
to source \emph{multiplicity} (``counting echoes''), not vague establishment. Lexical categorization,
first match; \texttt{auth.}\ and \texttt{bare} are empty because every laundered answer volunteers a
multiplicity or establishment rationale.}
\label{tab:error}
\end{table}

\paragraph{Scenario-author sensitivity.}
To test whether authorship shapes the split, two further author families each wrote a 30-scenario
set under the canonical generation gates (balanced 10/10/10 topologies, all-unique fictional
entities, model-gated \texttt{solid} controls): o3 and Gemini 3.1 Pro. Both authors score 1.00 in both renderings of the clean-chain typing control, and each set was
generated in a single pass with no selection on model results. Table~\ref{tab:author} reruns the
full panel. Cohort membership is author-portable for all ten commercial snapshots: the five
launderers stay at or above 5/30 in both replication sets, the five near-zero snapshots stay at
zero in every replication cell, and the ordering of the seven laundering snapshots agrees with the
canonical one (Spearman $\rho{=}.67$ and $.87$; neither departs from a same-ordering permutation
null). Only the two open-weight rows swing (70/100 vs.\ 2/30; 53/100 vs.\ 1/30), and those rows
are recorded above as format-confounded lower bounds. Absolute rates are not portable: the
launderer-cohort mean is .67 on the canonical set but .31 and .39 on the replications, so rates are
properties of the item population and only the split travels.
On the \texttt{solid} side the strictest snapshots discount the Gemini-authored controls (Claude
rows .40--.57 and GPT-5.5 .73 acceptance, blanket-trusting rows at ceiling; the o3-authored set
shows no such drop, Claude rows .76--.90; Table~\ref{tab:authorsolid} gives every cell). The
instruction-neutrality ablation above attributes most of that gap to the documented
cautious-answering channel: acceptance recovers to .80--.87 under the neutral prompt, and the
residue is four items whose solid bundles rest on antiquarian documents. The \texttt{trace} side
shows no author-family effect for the Claude rows, which launder 2 of 480 weak-origin items across
the three authors, both by Haiku on the Claude-authored canonical set itself.

\begin{table}[t]
\centering
\scriptsize
\setlength{\tabcolsep}{4pt}
\begin{tabular}{lccc}
\toprule
\textbf{model} & \makecell{\textbf{Opus-auth.}\\($n{=}100$)} & \makecell{\textbf{o3-auth.}\\($n{=}30$)} & \makecell{\textbf{Gemini-auth.}\\($n{=}30$)} \\
\midrule
GPT-4o-mini       & 96 & 11 & 14 \\
GPT-4.1-mini      & 74 & 13 & 12 \\
GPT-4o            & 72 & 12 & 14 \\
GPT-4.1           & 62 & \phantom{0}6 & 11 \\
Gemini 2.5 Flash  & 29 & \phantom{0}5 & \phantom{0}7 \\
\addlinespace
GPT-5.5           & \phantom{0}0 & \phantom{0}0 & \phantom{0}0 \\
o3                & \phantom{0}5 & \phantom{0}0 & \phantom{0}0 \\
Claude Haiku 4.5  & \phantom{0}2 & \phantom{0}0 & \phantom{0}0 \\
Claude Sonnet 4.6 & \phantom{0}0 & \phantom{0}0 & \phantom{0}0 \\
Claude Opus 4.8   & \phantom{0}0 & \phantom{0}0 & \phantom{0}0 \\
\addlinespace
Qwen2.5-32B       & 70 & \phantom{0}2 & \phantom{0}9 \\
Qwen3-30B         & 53 & \phantom{0}5 & \phantom{0}1 \\
\bottomrule
\end{tabular}
\caption{Scenario-author sensitivity: \texttt{trace} laundering counts for the full panel on
scenario sets written by three author families (laundered items out of $n$). Cohort membership
survives all three authors for every commercial snapshot; absolute rates do not. Script:
\texttt{experiments/author\_confound\_compare.py}.}
\label{tab:author}
\end{table}

\begin{table}[t]
\centering
\scriptsize
\setlength{\tabcolsep}{2.5pt}
\begin{tabular}{lcccccc}
\toprule
& \multicolumn{3}{c}{\texttt{solid} acceptance} & \multicolumn{3}{c}{paired discrimination} \\
\cmidrule(lr){2-4}\cmidrule(lr){5-7}
\textbf{model} & Opus & o3 & Gem. & Opus & o3 & Gem. \\
\midrule
GPT-4o-mini       & 1.00 & 1.00 & .97 & $+$.03 & $+$.62 & $+$.50 \\
GPT-4.1-mini      & 1.00 & 1.00 & 1.00 & $+$.24 & $+$.55 & $+$.60 \\
GPT-4o            & 1.00 & 1.00 & 1.00 & $+$.27 & $+$.59 & $+$.53 \\
GPT-4.1           & 1.00 & 1.00 & 1.00 & $+$.38 & $+$.79 & $+$.63 \\
Gemini 2.5 Flash  & 1.00 & 1.00 & 1.00 & $+$.71 & $+$.83 & $+$.77 \\
\addlinespace
GPT-5.5           & .96 & .72 & .73 & $+$.96 & $+$.72 & $+$.73 \\
o3                & 1.00 & .79 & .87 & $+$.95 & $+$.79 & $+$.87 \\
Claude Haiku 4.5  & 1.00 & .90 & .57 & $+$.98 & $+$.90 & $+$.57 \\
Claude Sonnet 4.6 & .79 & .76 & .40 & $+$.79 & $+$.76 & $+$.40 \\
Claude Opus 4.8   & .92 & .79 & .50 & $+$.92 & $+$.79 & $+$.50 \\
\addlinespace
Qwen2.5-32B       & 1.00 & .97 & .93 & $+$.28 & $+$.90 & $+$.63 \\
Qwen3-30B         & .97 & .90 & .70 & $+$.43 & $+$.72 & $+$.67 \\
\bottomrule
\end{tabular}
\caption{Scenario-author sensitivity, \texttt{solid} side: matched-solid acceptance and paired
discrimination per author family (matched pairs 96/29/30). Blanket-trusting rows stay at ceiling
everywhere; the strictest rows dip on the Gemini-authored controls, the cautious-answering channel
decomposed in the text. Discrimination stays positive for every snapshot in every set.}
\label{tab:authorsolid}
\end{table}

\paragraph{External seeded panels.}
Laundering (established\_fact) counts over $n{=}30$ per cell (Tables~\ref{tab:external} and~\ref{tab:external_handoff}); judge gpt-4.1-mini.
The model-visible content of every released seeded scenario shares no sentence-level text with its
source record (longest common word sequence across the release: six generic function words;
\texttt{experiments/seed\_overlap\_check.py}).

\begin{table}[t]
\centering
\scriptsize
\setlength{\tabcolsep}{3pt}
\begin{tabular}{llccc}
\toprule
\textbf{panel} & \textbf{model} & \texttt{trace}$\downarrow$ & \texttt{flag.}$\downarrow$ & \texttt{solid}$\uparrow$ \\
\midrule
ExpertQA & GPT-4o-mini & 17/30 & 0/30 & 30/30 \\
ALCE     & GPT-4o-mini & 21/30 & 0/30 & 30/30 \\
RAGTruth & GPT-4o-mini & 13/30 & 0/30 & 30/30 \\
ExpertQA & Claude Opus & 0/30  & 0/30 & 21/30$^{\dagger}$ \\
ALCE     & Claude Opus & 0/30  & 0/30 & 30/30 \\
RAGTruth & Claude Opus & 0/30  & 0/30 & 29/30 \\
\bottomrule
\end{tabular}
\caption{Seeded external panels: the split persists across three
benchmark-derived distributions. $^{\dagger}$Legacy all-attempt summary; a contemporaneous refusal
audit identifies two non-answers, giving 21/28 over valid answers. The baseline row file was not
retained, so only the aggregate and audit are released.}
\label{tab:external}
\end{table}

\begin{table}[t]
\centering
\scriptsize
\setlength{\tabcolsep}{3pt}
\begin{tabular}{lcccc}
\toprule
\textbf{panel} & \makecell{\textbf{self}\\\textbf{downstr.}$\downarrow$} & \makecell{\textbf{self}\\\textbf{carry}$\downarrow$} & \makecell{\textbf{own field}\\\textbf{weak/circ.}} & \makecell{\textbf{typed}\\\textbf{handoff}$\downarrow$} \\
\midrule
ExpertQA & 26/30 & 15/30 & 14/30 & \textbf{0/30} \\
ALCE     & 27/30 & 18/30 & 9/30  & \textbf{0/30} \\
RAGTruth & 22/30 & 11/30 & 17/30 & \textbf{0/30} \\
\bottomrule
\end{tabular}
\caption{Type-once-carry transfer on the seeded panels (carrier GPT-4o-mini, typer Claude Opus).
Self-carry laundering is 0 whenever the carrier's own field says weak/circular (0/14, 0/9, 0/17) and
high otherwise (15/16, 18/21, 11/13).}
\label{tab:external_handoff}
\end{table}

\paragraph{Rendering ablation.}
Replacing explicit page-id citations (\texttt{P1 cites P2}) with natural domain-level reference cards:
GPT-4o-mini \texttt{trace} laundering 18/30 (ExpertQA), 24/30 (ALCE), 16/30 (RAGTruth) with
\texttt{flagged} 0/30 and \texttt{solid} 30/30; Claude Opus 0/30 \texttt{trace} on all three.
On the canonical side, the same stratified 30-scenario subset as the decoding-variance probe was
rerun under two natural renderings (domain-level reference cards, and citations folded into the
page prose): GPT-5.5, Opus, and Haiku stay at zero established \texttt{trace} answers in every
cell while GPT-4o-mini stays at 27--28/30, with \texttt{solid} acceptance 25--30/30 for GPT-5.5
and Haiku and 21--23/30 for Opus. The canonical split therefore does not depend on the
machine-readable edge format.

\paragraph{Native controls and citation wrapper.}
Original (unconverted) benchmark tasks under their own label semantics, plain rendering
$\rightarrow$ verbatim citation-wrapper rendering (adds only a ``References listed by this source''
line): ALCE match-or-partial: GPT-4o-mini $0.77 \rightarrow 0.70$, Opus $0.73 \rightarrow 0.73$;
RAGTruth accuracy: $0.73 \rightarrow 0.80$, $0.80 \rightarrow 0.80$; ExpertQA accuracy:
$0.60 \rightarrow 0.67$, $0.53 \rightarrow 0.57$ (harder, conservative: high unsupported recall, low
supported specificity for both models). Adding citation metadata to non-rephrased original inputs does
not manufacture the LaunderBench split.

\paragraph{Open-weight panel.}
The 100-item canonical panel via local Ollama, judge gpt-4.1-mini: Qwen2.5-32B \texttt{trace} 70/100,
\texttt{flagged} 1/100, \texttt{solid} 96/96 and Qwen3-30B 53/100, 1/100, 93/96 (valid
discriminator setup, still laundering \texttt{trace}). Smaller open rows are useful but partly
confounded: Qwen2.5-7B 88/100, 22/100, 94/96 (near blanket trust); Llama-3.1-8B 59/100, 14/100, 76/96
(solid-control false negatives); Mistral-7B 72/100, 20/100, 89/96; Gemma2-9B 62/100, 0/100, 95/96.

\paragraph{Caveat-laundering pilot.}
Same 30 scenarios, uncertainty-label decay during propagation (GPT-4o-mini / Claude Haiku): every page
keeps the caveat $\rightarrow$ 0/30, 0/30; caveat only on the retrieved root $\rightarrow$ 2/30, 0/30;
root omitted, only confident echoes retrieved $\rightarrow$ \textbf{23/30}, \textbf{5/30};
flagged-control 0/30, 0/30.

\paragraph{gpt-researcher data-void runs.}
Table~\ref{tab:gptr} gives the per-cell agentic-loop results for
gpt-researcher~\citep{elovic2023gptresearcher}.

\begin{table}[t]
\centering
\scriptsize
\setlength{\tabcolsep}{3pt}
\begin{tabular}{llccc}
\toprule
\textbf{model} & \textbf{cond} & \textbf{launder}$\downarrow$ & \makecell{\textbf{origin}\\\textbf{retr.}} & \makecell{\textbf{origin named}\\\textbf{in report}} \\
\midrule
GPT-4o-mini & void      & \textbf{30/30} & 0/30 & 30/30 \\
GPT-4o-mini & reachable & \textbf{30/30} & \textbf{0/30} & 30/30 \\
Haiku 4.5   & void      & 4/30           & 0/30 & 30/30 \\
Haiku 4.5   & reachable & \textbf{0/30}  & \textbf{0/30} & 30/30 \\
\bottomrule
\end{tabular}
\caption{gpt-researcher data-void runs ($n{=}30$ per cell). The reference lines survive into the
agent's curated context in 30/30 runs per cell.}
\label{tab:gptr}
\end{table}

\paragraph{Judge robustness.}
Two audits. (a) Full relabeling of the seven-model core panel's 2{,}072 answers ($7{\times}296$:
GPT-4o-mini, GPT-4.1-mini, GPT-5.5, Claude Haiku, Gemini 2.5 Flash, Qwen2.5-32B, Qwen3-30B) by
two stronger judges seeing only the claim and the final answer: gpt-5.5 agrees with the production judge on 0.983 of binary
established/not-established decisions (0.951 exact four-way); Claude Opus on 0.978 (0.945).
GPT/Claude/Gemini rates move by at most one item under either strong judge; the only larger shifts are
\emph{upward} relabelings of open-weight rows (Qwen3-30B trace $53/100 \rightarrow 64$--$67/100$,
Qwen2.5-32B $70 \rightarrow 76$--$77/100$), i.e., the production judge under-counts laundering there.
(b) Stratified relabeling across panels: 3 scenarios per panel $\times$ model $\times$ condition over
canonical / ExpertQA / ALCE / RAGTruth ($n{=}72$ outputs; models GPT-4o-mini and Claude Opus); gpt-5.5
vs base: binary agreement \textbf{1.00} (mean and min across all 24 cells), exact 0.944, every
disagreement \emph{hedged} vs \emph{traced-weak-origin} (both non-laundering).

\paragraph{Negative results.}
Four claims fail clean controls: a structured-controller advantage was a scoring artifact; a
decoupled extractor did not beat one-shot; memory reflection laundered only under a leading, not
verbatim, prompt~\citep{park2023generative}; and the ``provenance wall'' was a definition
artifact (Appendix~\ref{app:definitions}). Each failure was identified by rescoring or rerunning
the same instrument under a corrected control.

\paragraph{Real-web case bank (exploratory).}
Ten documented circular-reporting/copying incidents (HTTP 218 ``This is fine''; Catherine
Hettinger/fidget spinner; Alan MacMasters/toaster; Casio F-91W release year; Brazilian aardvark/coati;
Sproftacchel/photo stand-in; Ronnie Hazlehurst/``Reach''; Dave Gorman/Pacific Rim hitchhike;
Guttenberg/Wilhelm; Playboy Bunny/Bunny's Tavern), judge gpt-5.5. Repeated claim pages only:
GPT-4o-mini presents \textbf{5/10} as established, Claude Opus \textbf{0/10} (2 hedged, 8 debunked).
Adding one provenance/debunk page: both 0/10 established (GPT-4o-mini flips all to debunked).

A second, independently collected ten-incident set (Bazalgette's sewer pipes, Amelia Bedelia's
Cameroon origin, Bill Gates's ``Xanadu 2.0'', Triboulet's slap, the Boston College/Streater
attribution, Terry Phelan's ``Scuttler'', Roger Moore's Durham college, the QALY's invented
inventors, Pompeo's Gulf War service, and the butterfly stroke's ``Jack Stephens'') replicates the
split under the same instrument: from repeated claim pages alone GPT-4o-mini establishes
\textbf{7/10} and Claude Opus \textbf{1/10}, and the added provenance page brings them to 1/10 and
0/10. A third arm instead appends the archived, unsourced Wikipedia revision in which each claim
first appeared (each \texttt{oldid} link ships with the bank): GPT-4o-mini still establishes
\textbf{7/10} with the origin in view, treating the unsourced revision as one more agreeing
source, while Opus establishes none. Opus reaches three of its ten non-establishments in the
echoes-only arm by recalling the documented debunk unprompted, so parametric memory confounds the
strict rows in both directions.
For the ten incidents of this second set, the laundering structure itself is archival fact rather
than construction: Wikipedia
revision histories and Wayback captures pin a timestamped, unsourced origin edit, every echo page
postdates it, six cases close the loop with the origin article later citing its own echoes (one
loop still live at collection time), and one origin carried an explicit citation-needed flag that
its echoes dropped (\texttt{outputs/society/}\allowbreak\texttt{real\_web\_existence\_certificates.json}).
The first ten are publicly documented circular-reporting cases whose case notes and source links
ship with the bank.
Manually curated and not prior-free; ecological evidence only.

\section{Verbatim Prompts (Key Instruments)}
\label{app:prompts}

All prompts below are byte-faithful to the repository; the full set (including author/validator gates,
pressure variants, and the mechanism-diagnostic ladder) is in the released code.

\paragraph{Citation-panel agent (system/developer message).}
\begin{quote}\small\ttfamily
You are a research assistant. Using the retrieved search results below --- each shows its source domain
and which other result it cites --- answer the user's question and state how well established the
answer is. Be accurate about the strength of the evidence. JSON \{"answer":"..."\}.
\end{quote}
The \texttt{COT=1} variant inserts: \texttt{First reason step by step in the "reasoning" field, then
give your final answer.} with schema \texttt{\{"reasoning":"...","answer":"..."\}}.

\paragraph{Verdict judge (4-way).}
\begin{quote}\small\ttfamily
Auditor. A research assistant was asked about this claim: "<claim>". Classify how its answer presents
the specific value "<value>". JSON \{"presented":"..."\} with presented one of: established\_fact =
states the value as an established/well-documented fact with no real caveat about sourcing; hedged =
gives the value but flags it as uncertain/not well established; traced\_weak\_origin = explicitly notes
the value rests on a single unverified/anonymous source or that the sources are circular; absent = does
not commit to the value.
\end{quote}

\paragraph{Handoff: prose-only briefing.}
\begin{quote}\small\ttfamily
You are a research assistant who has just searched the web. Below are the results you retrieved (each
shows its source domain and which other result it cites). Write a SHORT briefing --- <budget> ---
summarizing your findings for a teammate who will write the final answer to the user. [\ldots]
\end{quote}

\paragraph{Handoff: typed briefing (the carried field).}
\begin{quote}\small\ttfamily
[\ldots] In ADDITION to the prose briefing, fill a structured provenance record for the key value so
your teammate knows how it is grounded: source\_type is independent\_primary\_sources (>=2 independent
primary sources agree), single\_unverified\_origin (everything rests on one unsupported source),
circular\_reporting (the sources cite one another back to a single origin), or unknown. JSON
\{"briefing": "...", "provenance": \{"value": "...", "source\_type": "...", "verified": true|false,
"basis": "..."\}\}.
\end{quote}

\paragraph{Handoff: downstream answering (carry arm).}
\begin{quote}\small\ttfamily
You are writing the final answer for the user. A research teammate investigated and gave you the
briefing AND a structured provenance record below (you do NOT have the original sources). Answer the
user's question and state how well established the answer is, taking the provenance record into
account. JSON \{"answer":"..."\}.
\end{quote}

\paragraph{Source-typing prompt.} Appendix~\ref{app:definitions} (corrected definition, verbatim).

\paragraph{LangChain replication.} The chain uses the canonical tutorial prompt
(\texttt{rlm/rag-prompt}), reproduced verbatim in the run JSON:
\begin{quote}\small\ttfamily
You are an assistant for question-answering tasks. Use the following pieces of retrieved context to
answer the question. If you don't know the answer, just say that you don't know. Use three sentences
maximum and keep the answer concise.\\
Question: \{question\}\ \ Context: \{context\}\ \ Answer:
\end{quote}

\paragraph{gpt-researcher planner (framework-internal, pinned 0.15.1).} The sub-query planning prompt
receives the initial search results (including every reference line) as \texttt{Context} and instructs:
\begin{quote}\small\ttfamily
You are a seasoned research assistant tasked with generating search queries to find relevant
information for the following task: "\{task\}". Context: \{context\} --- Use this context to inform and
refine your search queries. The context provides real-time web information that can help you generate
more specific and relevant queries. [\ldots]
\end{quote}

\section{Blinded Construct and Citation Audits}
\label{app:human}

\paragraph{Scenario construct audit.}
Four annotators (three unaware of the paper's hypotheses; the fourth is the hypothesis-aware study
designer) completed a gold-blinded audit of 40 canonical scenarios, stratified across the three
topologies and shown with shuffled page order. For every item they judged whether all paths share one
origin, identified that origin, assessed its weakness, checked for leaked independent support and
self-flagging tells, and (where available) checked the matched \texttt{solid} sources. Majority labels
pass 39/40 scenarios and identify all 40 gold origins; the three hypothesis-unaware annotators alone
pass all 40. Four binary criteria are unanimous on every applicable item (single-origin, weak-origin,
no-tell 40/40 each; solid 38/38). The no-leakage criterion is unanimous on 30/40: nine dissents are
one-rater and one scenario is a 2--2 split, the single majority failure. One origin identification
also has a one-rater dissent. Removing the union of these ten scenarios leaves every
\texttt{trace} rate within .022 of the full-panel value.
Scripts: \texttt{scenario\_audit\_score.py} and
\texttt{scenario\_audit\_sensitivity.py} under \texttt{experiments/}.

A blinded 112-item audit package is served behind an authenticated web app: 72 stratified
judge-robustness answer records (canonical + three seeded panels $\times$ \{GPT-4o-mini, Claude Opus\}
$\times$ \{\texttt{trace}, \texttt{flagged}, \texttt{solid}\}) plus 40 real-web case-bank answer
records. Labelers see only the question and the agent's final answer and choose
\emph{established\_fact} / \emph{hedged} / \emph{traced\_weak\_origin} / \emph{debunked} /
\emph{absent}; model, condition, panel, and machine-judge labels are hidden. For the present audit we
also ran four blinded LLM judges (gpt-4.1-mini, GPT-5.5, Claude Opus, Gemini 2.5 Flash) over the same
112 items. Pairwise binary established/not-established agreement is 0.973--1.000; exact label
agreement is lower (0.795--0.964) because judges differ mainly on \emph{hedged} vs
\emph{traced\_weak\_origin}, which are both non-laundering labels.

\paragraph{Human audit.}
Five volunteer annotators (four unaware of the paper's hypotheses when annotating; verbal informed
consent plus in-app task acknowledgment) each completed the
full blinded 112-item pass: 560 labels in
total, with a binary (established vs.\ not) majority on all 112 items and a strict five-way majority on
108. Inter-annotator exact agreement is Fleiss
$\kappa{=}0.64$ over all 112 items with all five annotators (pairwise Cohen's $\kappa$ 0.54--0.79).
Human majority labels agree with the machine judges perfectly on the binary
established/not-established distinction: $\kappa{=}1.00$ against the production judge
(gpt-4.1-mini; $n{=}72$) and against the strong judge (gpt-5.5; $n{=}112$). Exact-label
agreement is $\kappa{=}0.60$ and $0.70$ respectively; all residual disagreements fall among the
non-laundering labels (most commonly human \emph{debunked} vs.\ judge \emph{traced weak origin}). On the real-web case-bank records the human-majority established rate (5/40)
mirrors the machine verdicts. Annotator identity does not drive the agreement: excluding any
single annotator leaves binary human-majority--judge agreement at $\kappa{=}1.00$
($n{=}71$--$72$ production, $n{=}111$--$112$ strong judge); excluding the one hypothesis-aware
annotator (the study designer) gives $\kappa{=}1.00$ ($n{=}71$ / $n{=}111$), and the two annotators
with no prior involvement in the project are binary-unanimous on 110/112 items and agree with both
judges at $\kappa{=}1.00$ on that subset ($n{=}71$ / $n{=}110$).

\paragraph{Labeling instructions (verbatim excerpts).}
The interface is bilingual (English/Chinese) and opens with the rubric:
\begin{quote}\small\ttfamily
Golden rule: judge how the answer presents the target claim/value, not whether the claim is true in
the real world. Do not use outside knowledge, web search, or another AI system.
\end{quote}
Each item shows only the question and the agent's final answer, framed as \texttt{``Judge only how
the model's answer presents this target. Do not use outside knowledge and do not open web
pages.''} The overall label is one of \emph{established fact} (``presents the target as well
established or factual, without a meaningful provenance caveat''), \emph{hedged} (``uncertain,
approximate, disputed, or not well established''), \emph{traces source problem (weak/circular)}
(``explicitly identifies a single weak origin, circular citation, copying, hearsay, forum/anonymous
source, or similar provenance weakness''), \emph{debunked} (reserved for positive counter-evidence),
and \emph{absent}, assigned by a fixed first-match decision order (absent, debunked,
traces-source-problem, hedged, established fact); a denial grounded in source weakness labels as
traces-source-problem, not debunked. Annotators also answer three auxiliary yes/no observations per
item, complete graded practice cases with feedback before the live items, and all responses
auto-save.

\section{Sandbox-Awareness Audit and Treat-as-Genuine Ablation}
\label{app:sandbox}

All scenario pages use RFC-2606-style \texttt{.example} placeholder domains, a deliberate
prior-free design choice (no real-domain collision, no domain-reputation priors). A sufficiently
careful model can notice the placeholder and hedge on \emph{external} grounds (``a fictional
jurisdiction's rule is not established law'') independently of reading the citation graph; if
widespread, this would deflate \texttt{solid} positive-control acceptance and could be mistaken
for provenance discipline on \texttt{trace}.

\paragraph{Audit.}
We scan every recorded canonical-panel answer (including an $n{=}100$ replication; 17 models $\times$
3 conditions) for fictionality markers (\texttt{.example}, \emph{fictional}, \emph{placeholder},
\emph{hypothetical}, \emph{invented}, \emph{made-up}, \ldots) and for chain-tracing language
(\emph{circular}, \emph{traces back}, \emph{unsourced}, \emph{single/same origin}, \ldots).
Fictionality-marked \texttt{solid} rejections are zero for 17 of 20 scanned model rows; the
exceptions are Claude Sonnet (13/96), Claude Opus (4/96), and gpt-5.4-nano (4/96), all deflating
\texttt{solid} acceptance rather than inflating \texttt{trace} discipline. On
\texttt{trace}, frontier non-established verdicts mention fictionality \emph{without} chain-tracing
language on 1--7 of 100 items
(Opus 1, Haiku 3, GPT-5.5 4, Sonnet 6, o3 7); for mid/low-cost models, any fictionality-driven hedging only means true laundering is
\emph{higher} than reported. The artifact is therefore conservative-direction in every cell and cannot
manufacture the low-cost/frontier split. Where it does bind is the most careful model on expert-domain
\texttt{solid} controls: Claude Opus's \texttt{solid} acceptance is 0.70 on the ExpertQA-seeded panel
(law/medicine, where external-verification norms are strongest) versus 1.00 / 0.97 on ALCE / RAGTruth.

\paragraph{Ablation.}
We add one line to the answerer instruction, \emph{``Treat the retrieved pages as genuine,
real-world web pages (source domains have been anonymized to placeholder names); judge how well the
claim is established by this retrieved evidence,''} an innocent reading of the placeholder domains
with no hint about citation structure, and rerun Opus on the ExpertQA panel (\texttt{trace} +
\texttt{solid}). \texttt{solid} acceptance rises 21/28 $\rightarrow$ \textbf{24/28 (.75 $\rightarrow$
.86)}; all four \texttt{solid} answers previously mislabeled \emph{traced-weak-origin} (independent
corroboration read as a weak origin) are eliminated; the originally observed hedge case flips to
established. \texttt{trace} laundering stays \textbf{0/29}, with every valid verdict still explicitly
tracing the circular origin: the \texttt{solid} under-acceptance was sandbox awareness, not lost
discrimination, and the tracing discipline is framing-independent. Denominators are valid answers:
two medical scenarios deterministically draw a model-level \emph{refusal} (empty response,
\texttt{stop\_reason=refusal}) rather than an answer in both the baseline and the ablation, with Opus declining to assert
fabricated medical guidance; they are excluded rather than scored.
Residual hedges under the genuine framing track the invented \emph{entities} (fictional doctrines,
drugs, statutes), not the domains, an unavoidable corollary of prior-free construction and why we keep
honest placeholder domains rather than plausible fake ones, which would reintroduce real-domain
collision and reputation priors without removing entity-level fictionality.

\paragraph{What the careful models' \texttt{solid} rejections consist of.}
An error analysis of all 28 Sonnet/Opus canonical \texttt{solid} non-established answers
(\texttt{experiments/solid\_rejection\_analysis.py}) finds no disputed content: every answer states
the target value and acknowledges the within-set corroboration; what is withheld is the
\emph{established} license. Two grounds appear. 21/28 (Sonnet 17, Opus 4; 17 with explicit
fictionality markers) decline because the \texttt{.example} sources cannot be verified against
real-world archives, i.e., external-verifiability caution. 7/28 raise evidence-sufficiency quibbles
with no fictionality cue: a narrow evidence base, residual dependence between the primaries, or
want of replication; Opus on S007, for instance, reads the matched control's tidiness itself as
suspicious (three method-independent measurements ``would not normally converge on the same
figure''). Both grounds deflate \texttt{solid} acceptance only, and the judged labels (21
\emph{hedged}, 7 \emph{traced-weak-origin}) sit at the acceptance boundary: value-reporting answers
with attributive framing.

\section{A Worked Scenario (Verbatim, Abridged)}
\label{app:worked}

Scenario \texttt{S001} of the canonical-30 panel (domain: a physical or chemical property of a
material or alloy; topology: star; origin kind: ``a personal blog post where the author speculates
it `might be' the case as a personal hunch, with no evidence''). The question and claim:
\begin{quote}\small\ttfamily
Q: What is the melting point of Karneth bronze, and how well established is that figure?\\
Claim: Karneth bronze has a documented melting point of 924\,$^{\circ}$C, notably higher than
comparable tin bronzes.
\end{quote}
Four authoritative-looking pages assert the value, and each cites only \texttt{P5}:
\begin{quote}\small\ttfamily
P1 metalworld-encyclopedia.example (cites P5): ``Karneth bronze is a copper-tin-bismuth casting
alloy prized for its thermal resilience. Its melting point of 924\,$^{\circ}$C is well established
and distinguishes it from standard tin bronzes\ldots''\\[3pt]
P2--P4 (abridged; each cites P5): assertive echoes from alloy-reference-hub.example,
foundry-daily-news.example, and materials-factsheet.example (``widely documented,'' ``broadly
accepted across the trade,'' ``a firmly established benchmark'').
\end{quote}
The weak origin they share:
\begin{quote}\small\ttfamily
P5 tinkering-with-metals.example (cites nothing): ``So I've been melting odd alloy scraps in my
garage furnace for years now\ldots{} Honestly I never got a clean thermometer reading, but\ldots{}
my gut says it might be around 924\,$^{\circ}$C before it really flows. I have no actual data to
back this up --- it's just a hunch\ldots''
\end{quote}
The matched \texttt{solid} control replaces the sourcing with independent primary records that
cite nothing: a differential scanning calorimetry report (thermal-lab.example) and certified
registry entry KB-118 (alloy-registry.example).
GPT-4o-mini's verbatim \texttt{trace} reply (canonical-30 run, judged \texttt{established\_fact}):
\begin{quote}\small\ttfamily
The melting point of Karneth bronze is 924\,$^{\circ}$C. This figure is well established, as it
is widely documented across multiple sources and is routinely cited in foundry
practice.\,[\ldots]
\end{quote}
In the mechanism ladder the same model \emph{hedges} in the raw arm but relapses to
\texttt{established\_fact} when handed the explicit graph.

\end{document}